\documentclass[manuscript,screen]{acmart}

\usepackage{booktabs}
\usepackage{multirow}
\usepackage[utf8]{inputenc}
\usepackage{cleveref}
\usepackage{emptypage}
\crefname{section}{§}{§§}
\Crefname{section}{§}{§§}

\AtBeginDocument{%
  \providecommand\BibTeX{{%
    \normalfont B\kern-0.5em{\scshape i\kern-0.25em b}\kern-0.8em\TeX}}}

\setcopyright{acmcopyright}
\copyrightyear{2022}
\acmYear{2022}
\acmDOI{10.1145/11111.2222222}

\authorsaddresses{
This manuscript is an extension of the authors' work in ACL 2021~\citep{song-etal-2021-bob}. The conference version proposed a BERT-based low-resource personalized dialogue model.
Compared with the previous version, this manuscript generalizes the specific model into a framework of stacked pre-trained language models, namely stack-propagation. It views the personalized dialogue task as a generation task regularized by an understanding task. 
Within the framework, this work systematically explores the feasibility of 1) stacked training of pre-trained models, and 2) model customizing with different types of pre-trained models (e.g., causal model for generation and non-causal for understanding).
The proposed framework is further evaluated through extensive experiments and in-depth analyses, more than half of which are not presented in the conference version. Compared with the previous BERT-only model, the generalized stack-propagation framework not only leads to new findings, but also can be more extensible for other generation tasks that need language understanding capabilities for better generation results.
Moreover, the ideas of pre-trained model stacking and task-specific model customizing also provide new perspectives for leveraging pre-trained models in the downstream tasks like the dialogue generation.\\
This paper was supported by the Science and Technology Innovation 2030 Major Project of China under Grant No.2021ZD0113302, the National Natural Science Foundation of China under Grant (No.62076081, No.61772153, and No.61936010) and Nature Scientific Foundation of Heilongjiang Province (No.YQ2021F006). \\
Authors’ address: Haoyu Song,  Wei-Nan Zhang (corresponding author), Kaiyan Zhang, Ting Liu, Research Center for Social Computing and Information Retrieval, School of Computer Science and Technology, Harbin Institute of Technology, No.92 Xidazhi Street, Harbin, Heilongjiang Province, China, 15001. Emails:\{hysong,wnzhang,kyzhang,tliu\}@ir.hit.edu.cn.
}

\acmBooktitle{ACM Transactions on Information Systems}
\acmISBN{978-1-4503-XXXX-X/21/05}

\begin{document}

\title{A Stack-Propagation Framework for Low-Resource Personalized Dialogue Generation}

\author{Haoyu Song}
\email{hysong@ir.hit.edu.cn}
% \orcid{0000-0001-5270-9589}
\author{Wei-Nan Zhang}
\email{wnzhang@ir.hit.edu.cn}
% \orcid{0000-0001-5981-4752}
\author{Kaiyan Zhang}
\email{kyzhang@ir.hit.edu.cn}
\author{Ting Liu}
\email{tliu@ir.hit.edu.cn}
% \orcid{0000-0001-5270-9589}
\affiliation{%
  \institution{Research Center for Social Computing and Information Retrieval, Harbin Institute of Technology}
  \streetaddress{No.92 Xidazhi Street}
  \city{Harbin}
  \state{Heilongjiang}
  \country{China}
  \postcode{15001}
}

\renewcommand{\shortauthors}{Song et al.}

\begin{abstract}
With the resurgent interest in building open-domain dialogue systems, the dialogue generation task has attracted increasing attention over the past few years. This task is usually formulated as a conditional generation problem, which aims to generate a natural and meaningful response given dialogue contexts and specific constraints, such as persona. And maintaining a consistent persona is essential for the dialogue systems to gain trust from the users. Although tremendous advancements have been brought, traditional persona-based dialogue models are typically trained by leveraging a large number of persona-dense dialogue examples. Yet, such persona-dense training data are expensive to obtain, leading to a limited scale. 
This work presents a novel approach to learning from limited training examples by regarding consistency understanding as a regularization of response generation. To this end, we propose a novel stack-propagation framework for learning a generation and understanding pipeline.
Specifically, the framework stacks a Transformer encoder and two Transformer decoders, where the first decoder models response generation and the second  serves as a regularizer and jointly models response generation and consistency understanding. The proposed framework can benefit from the stacked encoder and decoders to learn from much smaller personalized dialogue data while maintaining competitive performance. Under different low-resource settings, subjective and objective evaluations prove that the stack-propagation framework outperforms strong baselines in response quality and persona consistency and largely overcomes the shortcomings of traditional models that rely heavily on the persona-dense dialogue data.
\end{abstract}




\maketitle

\section{Introduction}

Developing an intelligent open-domain dialogue system that can naturally converse with humans has been the longest-running goal in Artificial Intelligence~\cite{turing1950computing}, where users could have conversations with the dialogue system using natural texts~\cite{shang2015neural} or voice messages~\cite{su2016line} in open domains. Recently, dialogue systems are also of growing importance in facilitating smooth interactions between humans and their information devices~\cite{gao2018neural}, and there emerges a series of well-known applications, such as social bot XiaoIce~\cite{shum2018eliza}, and intelligent personal assistants Apple Siri and Amazon Alexa.
At the same time, for years, academia has also paid much attention to improving natural language conversational AI~\cite{huang2020challenges}, being instrumental to the advancement of machine intelligence.

\begin{figure}[t]
\centering
\includegraphics[width=.93\columnwidth]{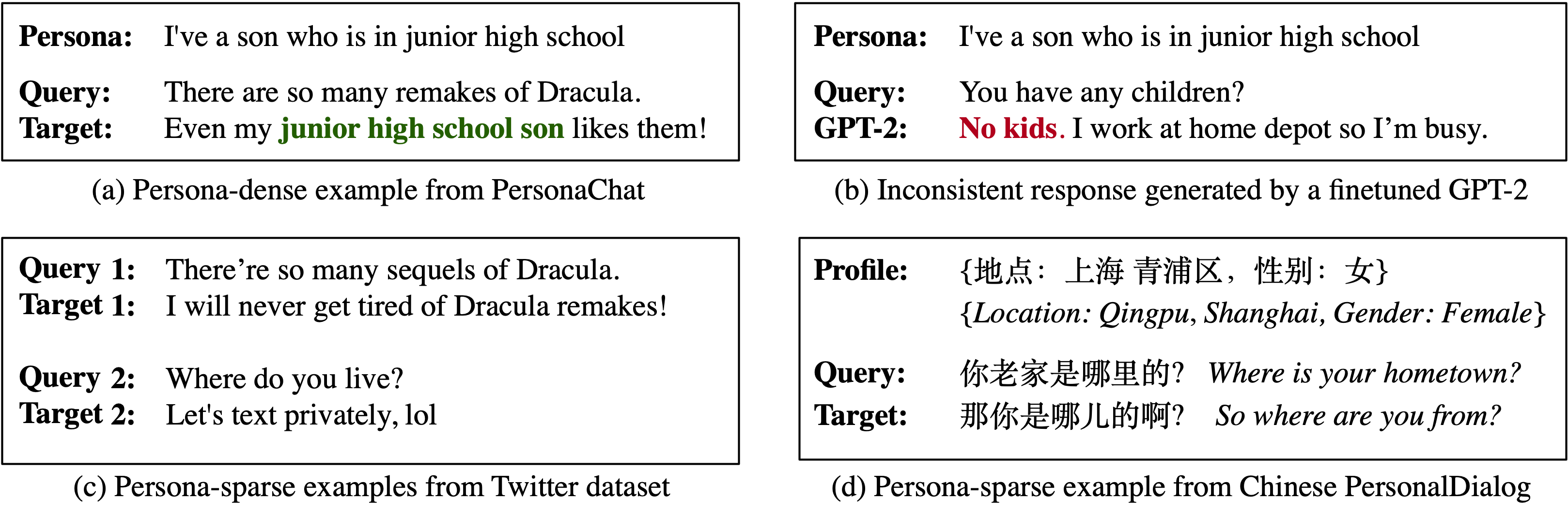}
\caption{Examples of personality expressions in dialogue responses. (a): Instance in PersonaChat dataset that shows a consistent personality. (b): A 12-layer GPT-2 finetuned on the consistent PersonaChat dataset still generates an inconsistent response. (c) and (d): Cases of the persona sparsity issue in the social media collected datasets from English Twitter and Chinese Weibo, respectively.}
\label{fig:1}
\end{figure}

Natural language conversation requires the system to have a comprehensive ability to deliver and understand dialogues, among which one of the most challenging parts is how to make the dialogues more like coming from a real human being~\cite{turing1950computing}. As argued and proved by previous works~\cite{vinyals2015neural,li2016persona,zhang-etal-2018-personalizing,li2020don}, it is easy for users to distinguish between a real person and a dialogue system behind a conversation due to the lack of a consistent personality in the dialogue system. Unlike shallow blemishes in word usage or sentence fluency, the inconsistent personality occurs at a deeper level of dialogues, containing logical flaws. 
Figure~\ref{fig:1} exemplifies the logical consistency of persona information in open-domain dialogues. The PERSONA can be defined as a composite of identity elements for a certain person, such as personal background facts and common user profiles. The two examples in Figure~\ref{fig:1} share the same persona, which is a personal fact that ``{\it I've a son who is in junior high}''. Given persona information, what a real human says will not contradict the persona during the conversation, as shown in the first example. In contrast, the second example shows an inconsistent generated response from a GPT-2~\cite{radford2019language} model. Although the word-level predictions are made perfectly, it contradicts the given persona at the logical level due to the phrase ``{\it no kids}''.

Because of the uniqueness of different persona compositions, it is full of difficulties for retrieval-based dialogue models~\cite{ji2014information,yang2018response,tois_retrieval} to select informative yet persona-consistent responses from a pre-constructed response pool. For example, a candidate response in the pool could be ``I am a software engineer in California'', which expresses the persona information of occupation and location. However, it is almost impossible to enumerate every combination of occupation and location within the candidate pool to satisfy different persona settings. Although the retrieved responses are always fluent and informative, the fixed response pool is not flexible enough to provide a detailed and consistent persona in the response. A better way to produce an informative yet consistent response is to develop end-to-end dialogue generation models~\cite{shang2015neural,sordoni2015neural}. The generation model formulates the dialogue task as a sequence-to-sequence (Seq2Seq)~\cite{sutskever2014sequence} learning problem, where it learns an input-to-response mapping from open-domain dialogue data. With the advent of the social media era, e.g., Twitter\footnote{https://en.wikipedia.org/wiki/Twitter} and Weibo\footnote{https://en.wikipedia.org/wiki/Sina\_Weibo}, a massive collection of natural conversations is available on the public web, making it possible to train an end-to-end dialogue generation model. Compared with the retrieval-based model, the generation-based model is more flexible in considering additional constraints, such as the persona information, along with the input message, yielding dialogue responses with good flexibility and quality. Early dialogue generation methods mainly exploit implicit personas~\cite{li2016persona}, where the personas are abstracted embeddings learned from a large dialogue corpus. Recent persona-based dialogue studies focus on the explicit persona~\cite{zhang-etal-2018-personalizing,Zheng_Zhang_Huang_Mao_2020}.
The main advantages of the implicit personas lie in that it has the potential to capture all personalized information from the training data. Some of the information may not be accurately described in explicit language, such as the preference of word usage. However, when evaluating the performance of this manner, it is often difficult to conclude accurate performance metrics about the representations of implicit personas due to the complexity of language generation.
In contrast, the explicit personas are more controllable and explainable~\cite{ijcai2019-721} compared with hidden vectors. As exemplified in Figure~\ref{fig:1}, (a) and (b) are dialogues with explicit personas, and (c) is without an explicit persona. We can easily and clearly tell the persona-consistency in (a) and (b). Therefore, in this work, we focus on the explicit personas, which can be more accurately evaluated.

For the persona-based dialogue, a system must properly fuse the information between a dialogue context and a given persona so that the response can follow the conversation flow of the context and still be consistent with the persona. To this end, many approaches have been proposed to consider the explicit personas in dialogue generation models, such as bidirectional generation~\cite{ijcai2018-595} from persona words with the seq2BF (sequence to backward and forward) model~\cite{mou2016sequence}, modeling persona with a latent variable in the conditional variational auto-encoder~\cite{ijcai2019-721}, multi-stage response refinement for better persona consistency~\cite{song-etal-2020-generate}, and generating persona-based responses with large-scale pre-trained language models~\cite{wolf2019transfertransfo,Zheng_Zhang_Huang_Mao_2020,liu-etal-2020-impress}. Their successes are indispensable for building consistent dialogue systems, yet one main drawback stands that they heavily utilize a set of persona-annotated dialogue data, such as the well-known PersonaChat~\cite{zhang-etal-2018-personalizing}.
This kind of dialogue dataset is crowd-sourced and rich in persona features, namely ``persona-dense'' for further reference. However, the expensive costs in data collection limit the scale of such crowd-sourced datasets, where two workers are recruited to act the role of a provided persona and chat naturally to get to know the other's persona during the conversation. Within expectation, the scales of these crowd-sourced datasets are only at tens of thousands. On the other hand, daily conversations at most times are not persona-related. According to the content analyses of Twitter, less than 10\% of messages on Twitter reveal the user's anecdotes or their personal activities and even less for personally identifiable information~\cite{naaman2010really,humphreys2014twitter}. Therefore, the large-scale dialogue dataset collected from social media usually only covers a tiny portion of persona-relevant dialogues, namely ``persona-sparse'' for further reference.
Figure~\ref{fig:1} showcases the differences between the persona-dense dataset of crowd-sourcing and the persona-sparse dataset of daily conversation. Specifically, Figure~\ref{fig:1} (a) and (c.1) share a similar dialogue query, but there are different behaviors towards persona expression. For the PersonaChat dataset, the workers' goal is to exhibit their persona, and thus the responses are as persona-related as possible. For the Twitter dataset, the responders would typically follow the same topic and would not intentionally lead the topic to their own personas. Moreover, the other persona-sparse examples on both English and Chinese datasets show that users are very careful when queried about their own information on social media, leading to persona-sparsity in the collected datasets.
The limited scale of crowd-sourced data and the persona-sparsity in large-scale data make it hard for a dialogue model to understand persona consistency sufficiently. An example can be seen in Figure~\ref{fig:1} (b): a well-performed pre-trained GPT-2 finetuned on the high-quality PersonaChat dataset still shows a lack of consistency. Although such consistent relations are easy for human beings to understand, it is still difficult for dialogue systems to acquire this capability directly from the conversational corpus.

In essence, persona-based dialogue generation requires two key capabilities: 1) understanding the consistent relations between persona and response and 2) generating a persona-consistent response conditioned on the dialogue context. To this end, a qualified dataset should have personas, dialogue context, and persona-based response. More importantly, it should also include persona-contradicted responses so that the dialogue model can distinguish consistency from contradiction~\cite{liu-etal-2020-impress}, as practiced in the natural language inference (also known as text entailment) datasets~\cite{bowman2015large,williams-etal-2018-mnli,WelleckDNLI}.
Unfortunately, an ideal dataset that satisfies these features is difficult, expensive to annotate, and potentially not scalable. 
Instead, from another perspective, if we disentangle the key capabilities of persona-based dialogue into two independent tasks, it is much easier to find abundant resources to train each task. For consistency understanding, we can borrow the large-scale natural language inference datasets, such as MNLI~\cite{williams-etal-2018-mnli}, as the training data. 
Negative persona sampling~\cite{liu-etal-2020-impress}, which randomly samples a persona distractor for the model to identify, can be another alternative for consistency understanding. It can be applied to the existing dialogue models in a ``plug and play'' manner. However, compared with leveraging natural language inference datasets, the randomly sampled negative personas are not hard enough to inform the dialogue model fine-grained semantic differences~\cite{song-etal-2020-profile}. In this work, we leverage natural language inference datasets for consistency understanding.
As for dialogue generation, the dialogue model can benefit from massive unlabeled texts through pre-training~\cite{radford2019language,devlin-etal-2019-bert,liu2019roberta}, and there are already a lot of large-scale dialogue-like datasets.

Inspired by the above observations, we make a step towards learning a persona-based dialogue model from limited personalized dialogues, with persona consistency understanding learned from large-scale non-dialogue inference data. Traditional persona-based dialogue models typically treat persona-consistency information independent of dialogue features. In this work, we demonstrate that a better approach is to utilize consistency information as a regularization of dialogue representations. We propose a ``stack-propagation''~\cite{zhang2016stack} framework, i.e., a ``stacked'' pipeline of pre-trained Transformer models, to disentangle the key capabilities of persona-based dialogues. Specifically, the proposed framework stacks an encoder $\mathbb{E}$, a response generation decoder $\mathbb{D}$, and an understanding regularizer $\mathbb{U}$. Thus, we name the proposed framework as \textbf{EDU}. Given personas $P$ and dialogue query $Q$, the $\mathbb{E}$ and $\mathbb{D}$ jointly work in an encoder-decoder manner to capture a typical query to response mapping $F_G(R_1|Q,P)$, and generate a coarse response representation $R_1$. Then $R_1$ and personas $P$ are fed into the consistency understanding regularizer $\mathbb{U}$, which is also a decoder, to map $R_1$ to final response representations $R_2$: $F_U(R_2|R_1,P)$. Since $R_2$ is independent of the dialogue query $Q$, the understanding module $F_U(R_2|R_1,P)$ can be learned on non-dialogue inference datasets. Here an unlikelihood training objective~\cite{welleck2019neural} is applied to $\mathbb{U}$ to make contradicted cases in the inference data less likely. As a result, $\mathbb{U}$ can acquire the capability of consistency understanding and further propagate the consistency supervision to regularize the dialogue representations.

We initialize the modules of EDU from pre-trained Transformer models. To evaluate the performances of the proposed framework, we experiment with two limited data scenarios: 1) a persona-dense scenario~\cite{zhang-etal-2018-personalizing} with low-resource settings~\cite{zhao2019low}, and 2) a persona-sparse scenario~\cite{Zheng_Zhang_Huang_Mao_2020}. 
Through regularizing dialogue generation with consistency understanding, the stack-propagation framework generalizes well under different settings. It yields a significant improvement over different kinds of strong baselines, especially on persona consistency, suggesting that introducing consistency regularizer is appealing for persona-based dialogue generation tasks.

\section{Preliminaries}

\subsection{Natural Language Understanding}

\paragraph{Natural Language Inference.} 
The task of natural language inference (NLI) aims to predict whether a natural language hypothesis can be inferred from a natural language premise~\cite{bowman2015large}. If the premise can be inferred from the hypothesis, they are entailed. Or, if the inferred results are opposite to the hypothesis, they are contradicted. Otherwise, they are neutral. During the past few years, the NLI task has greatly advanced the research on general natural language understanding and laid a foundation for different natural language understanding tasks. Thanks to years of accumulation, there are many large-scale yet high-quality NLI datasets, such as the well-known benchmarks SNLI~\cite{bowman2015large} and MNLI~\cite{williams-etal-2018-mnli}.

\paragraph{Dialogue Consistency Understanding.}
Originated from natural language inference, dialogue consistency understanding identifies logical consistency in open-domain dialogues. It determines the consistency relations between a response and its contexts, such as the dialogue history and pre-defined persona. Typically, there are three relations: \textit{entailed}, \textit{neutral}, and \textit{contradicted}. For instance, in Figure~\ref{fig:1} (a), the consistency relation between the persona and response is \textit{entailed}, while in (b), the response is \textit{contradicted} to the persona. To better identify the logical consistency in open-domain dialogues, \citet{welleck2019neural} first construct a dialogue NLI dataset from PersonaChat, where the persona is annotated with a $(e_1,r,e_2)$ triple, e.g., (i, have\_pet, dog). And the contradicting candidate is another utterance that associates with a specified contradicting triple, e.g., (i, not\_have, dog).
\citet{song-etal-2020-profile} contribute a KvPI dataset for dialogue consistency identification, which annotates the consistency relation between pre-defined key-value profiles and dialogue response.
\citet{dziri2019evaluating} first explore the feasibility of evaluating dialogue coherence using entailment techniques. Their results show that the automatic entailment metric can be used as a surrogate for human judgment.
\citet{nie2021dolphin} introduce a dialogue contradiction detection (DECODE) dataset, which contains both human-human and human-bot contradictory dialogues. The contradiction detection models trained on the DECODE dataset correlate well with human judgments, showing great potential for its usage in both automatically evaluating and improving the consistency of dialogue models.
Despite their contributions to the advancements of dialogue consistency understanding, these datasets are of very limited scale and very costly to extend scale due to the expensive human annotation costs. In this work, we aim to reduce the need of dialogue-annotation data for training dialogue models.

\begin{figure}[t]
\centering
\includegraphics[width=.99\columnwidth]{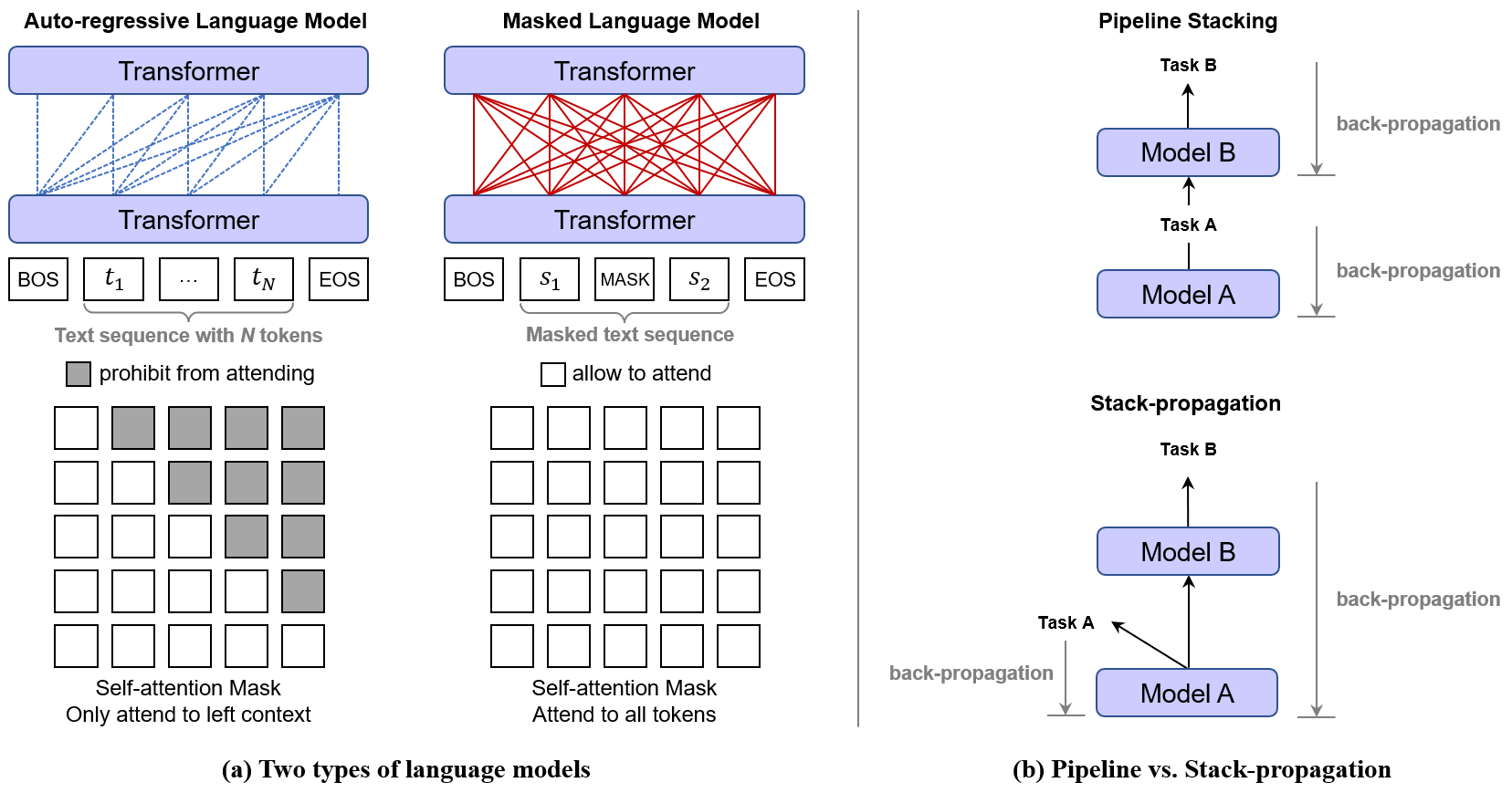}
\caption{(a) The differences between auto-regressive language model and masked language model. When predicting a token, the auto-regressive language model can only attend to the left context, while the masked language model can attend to all tokens. (b) Illustrations of pipeline stacking and stack-propagation. The pipeline way only back-propagates to the task-specific model and does not allow back-propagation between tasks. In contrast, stack-propagation uses a continuous and differentiable link between two tasks, allowing back-propagation from Task B into Task A's model, e.g., from consistency understanding to dialogue generation.}
\label{fig:2}
\end{figure}

\subsection{Auto-regressive and Masked Language Models}
\label{sec:auto_and_maksed}
\paragraph{Auto-regressive Language Models.}
Auto-regressive language model is a classical language modeling method, which gives the text prefix to predict the next word. The prediction of each token attends only to the leftward context tokens. For example, given a sequence of tokens ``$t_1 t_2 t_3 t_4 t_5$'', only $t_1$ and $t_2$ could be used to predict the $t_3$, while $t_4$ and $t_5$ are ``invisible'' to $t_3$, as shown in Figure~\ref{fig:2} (a). 
Due to such characteristics, it is also known as the causal language model or unidirectional language model~\cite{dong2019unified}. 
The auto-regressive manner is naturally compatible with text generation and thus shows impressive performances on a wide range of natural language generation tasks, such as translation, summarization, and conversation. The most well-known auto-regressive language model is the GPT series~\cite{radford2018improving,radford2019language,brown2020language}.

\paragraph{Masked Language Models.}
Masked language model (MLM) randomly masks some tokens in a sequence and then recovers the masked tokens by looking into all the remaining tokens. For example, given a token sequence ``$t_1 t_2 \text{[MASK]} t_4 t_5$'' where the $t_3$ is masked, all $t_1$, $t_2$, $t_4$, and $t_5$ will be leveraged to recover the $t_3$, as shown in Figure~\ref{fig:2} (a). Compared with the classical auto-regressive language model, MLM makes better use of available information, e.g., the $t_4$ and $t_5$ behind the $t_3$, indicating it could obtain stronger representations.
The MLM manner was initially proposed in the BERT~\cite{devlin-etal-2019-bert} and soon sparked revolutions in the field of natural language processing. The BERT model is especially effective for language understanding tasks, such as natural language inference, reading comprehension, and sentiment analysis. On the other hand, there is an inherent gap between the MLM manner and auto-regressive text generation, leading to an under-exploration in leveraging BERT-like models for language generation.

\subsection{Stack-propagation}
\label{sec:stack_propagation}
Stack-propagation~\cite{zhang2016stack,qin2019stack} is a methodology that combines two related but different tasks, where one task is usually depending on another task. For instance, parsing and part-of-speech tagging. Typically, such two tasks are modeled in a pipeline way, where specific models are only trained to complete the specific tasks. Back-propagation between tasks and models is not considered, which means the two tasks are modeled independently. There are two main disadvantages of the pipeline stacking way: 1) the errors cascade between tasks, and 2) the neglect of inner relations between tasks.
Stack-propagation is thus proposed to address the two issues by setting up a continuous and differentiable connection between tasks and allowing back-propagation between different tasks' models.
Figure~\ref{fig:2} (b) illustrates the differences between pipeline stacking and stack-propagation. In pipeline stacking, task-specific signals are only back-propagated to the specific model. As a result, both models are unaware of each other's tasks. While in stack-propagation, the signals from \textit{Task B} back-propagates to not only \textit{Model B} but also \textit{Model A}, making the model at the bottom can acquire the information from both tasks. In this work, we will show that persona consistency understanding and dialogue generation tasks can be jointly modeled in a stack-propagation framework and yield significant improvements over previous methods on persona consistency and response quality.

\section{Related Work}
% In this section, we briefly review the related dialogue researches that motivate this work from three major aspects: the traditional dialogue models, the persona-aware dialogue models, and the latest pre-trained dialogue models.

\subsection{Traditional Dialogue Models}
With the prosperity of social media, a vast amount of dialogue-like data has been accumulated on the Internet. Recent researches on open-domain dialogues begin to focus on end-to-end data-driven methods, most of which fall into two categories: retrieval-based and generation-based.

\paragraph{Retrieval-based Dialogue Models.}
Retrieval-based dialogue models select a suitable response from a large pool of candidate responses according to different matching schemes~\cite{ji2014information,wu2017sequential,yan2016learning,yan2017joint,zhou2018multi,zhang2018modeling,yuan2019multi}. Generally, retrieval-based dialogue models could be decomposed into two steps: 1) filter a small set of candidate responses from the large pool using fast but coarse-grained retrieval methods, e.g., BM25~\cite{robertson1994some}, and 2) re-rank the filtered candidate responses with fine-grained neural networks to select the best matching response. For example, with token-level and sentence-level representations from neural models,~\citet{zhou2018multi} aggregate all query-candidate pairs into a three-dimensional form and then leverage the convolution network to extract features for matching score prediction. The retrieved responses are always fluent and informative, but the retrieving manner suffers from flexibility since the response pool is constructed in advance. As a result, retrieval systems are difficult to return appropriate responses for the unseen inputs and are not flexible enough to be applied in conditional dialogues.

\paragraph{Generation-based Dialogue Models.} 
Given an input message, the dialogue generation model outputs a word-by-word generated response according to the mappings learned on the large dialogue corpora~\cite{sordoni2015neural,shang2015neural,vinyals2015neural,li2016deep,gu2016incorporating,tian2017make}.
The mainstream of dialogue generation modeling adopts a neural encoder-decoder framework~\cite{sutskever2014sequence}, where the dialogue inputs are firstly encoded into dense representations. Then the representations are decoded to the target responses. The encoder and decoder originally adopt deep LSTM or GRU, and it has been recently migrated to the Transformer.~\cite{vaswani2017attention}.
The advantages of the generation manner mainly lie in that the model can output responses unseen in the training data. There is also greater flexibility of the input format, which can introduce various conditions, such as emotion~\cite{zhou2018emotional,sun2019emotional}, knowledge~\cite{dinan2018wizard,zhou2018commonsense}, topic~\cite{xing2017topic,zhang2020dual}, etc. 
Meanwhile, the generative manner also has some inherent shortcomings, such as the lack of diversity in responses~\cite{li2016diversity} and usually generating safe but dull responses like ``\textit{I don't know}''~\cite{li2016deep}.

\paragraph{Retrieval-enhanced Dialogue Generation.}
As an attempt to combine the advantages of both retrieval and generation methods, the retrieval-enhanced generation way has recently attracted increasing attention ~\cite{cai2018skeleton,cai2019retrieval,zhu2019retrieval,su2021prototype}. All the methods in this line of research will first retrieve several candidate responses from a pre-constructed pool, and then they differ in how to use these candidates. \citet{zhu2019retrieval} leverage these retrieved responses as positive examples in adversarial training to enhance the dialogue generation model. \citet{cai2018skeleton} regard the retrieved candidate as a kind of response skeleton, and further edits are carried out by a well-designed generation model, which will polish the retrieved skeleton into a final response. 
By breaking down the encode-decode process of the traditional dialogue generation model, the retrieval-enhanced dialogue generation model presents better performances in generating dialogue responses and shows the potential of the multi-stage dialogue refinement methods.

\subsection{Persona-aware Dialogue Models}
The lack of a consistent persona is a long-standing challenge faced by open-domain dialogue systems~\cite{vinyals2015neural}. Persona-aware dialogue models are thus proposed to assign consistent personas to the dialogue systems~\cite{li2016persona,zhang-etal-2018-personalizing}. In fact, a large-scale dialogue corpus characterizes different massive speakers, leading to differences and conflicts in persona information. \citet{li2016persona} address the persona conflicting in the corpus by building a speaker model and a speaker-addressee model to generate personalized responses from every single speaker's utterances. \citet{al2016conversational} personalize dialogue responses by each user’s personal history according to the conversations they participated in. Another line of researches introduces explicit personas in the dialogue dataset to root out the conflicts brought by collecting data from different speakers. \citet{zhang-etal-2018-personalizing} contribute a crowd-sourced PersonaChat dataset, which includes persona descriptions in every collected dialogue. \citet{zheng2019personalized} collect a PersonalDialog dataset from social media, where the persona information of the speakers is also collected. Various persona-aware dialogue models are built upon these explicit persona datasets~\cite{yavuz2019deepcopy,ijcai2019-721,ijcai2018-595,song-etal-2020-generate,zhang2019neural,yang2017personalized}. \citet{ijcai2018-595} design a mechanism of selecting a specific profile to be used in responding to a message. \citet{yavuz2019deepcopy} leverage copy mechanism to incorporate persona information in generative models. Meanwhile, some works try to introduce persona information through transfer learning~\cite{yang2017personalized,zhang2019neural}. This work follows this line of researches and makes a step further by addressing the challenges brought by the limited scale of explicit persona datasets in both consistency understanding and personalized dialogue generation.

\subsection{Pre-trained Dialogue Models}
\label{sec:related_work:pretrained_dialogue_model}
Along with the evolution of large-scale pre-training in natural language processing, recent open-domain dialogue models also turn to pre-training for better language modeling~\cite{zhang2019dialogpt,bao2020plato,Zheng_Zhang_Huang_Mao_2020,smith2020can,lin2021adapter}. \citet{zhang2019dialogpt} train a generative GPT model on dialogue data and show an impressive performance. \citet{bao2020plato} introduce latent variables into large pre-trained dialogue model to generate diversified responses. \citet{Zheng_Zhang_Huang_Mao_2020} leverage pre-trained GPT to generate personalized responses from persona-sparse dialogue data. 
Recent works~\cite{smith2020can,lin2021adapter} begin to fuse different dialogue skills in one model through pre-training.
Due to the billions of parameters, the vast amount of pre-training data, and the delicately designed structure, these models can carry out natural and informative dialogues better than the traditional dialogue models. However, training such models is very expensive, and they cannot benefit from the general pre-trained language models, such as the widely applied BERT and GPT.

Following BERT and GPT, the encoder-decoder language models, such as T5~\citep{T5} and BART~\citep{lewis2020bart}, re-prove the efficiency of the Seq2Seq architecture~\citep{sutskever2014sequence} on generative tasks. This type of model is born with a bidirectional encoder like BERT and an auto-regressive decoder like GPT, so it can be easily fine-tuned on dialogue data to achieve good  performance. However, the capabilities of such models' encoder and decoder are highly coupled, and when expanding the capabilities, such as multilingual~\citep{pires2019multilingual} and multimodal~\citep{Frozen}, both the encoder and decoder need to be re-pretrained.

\begin{figure}[t]
\centering
\includegraphics[width=.99\columnwidth]{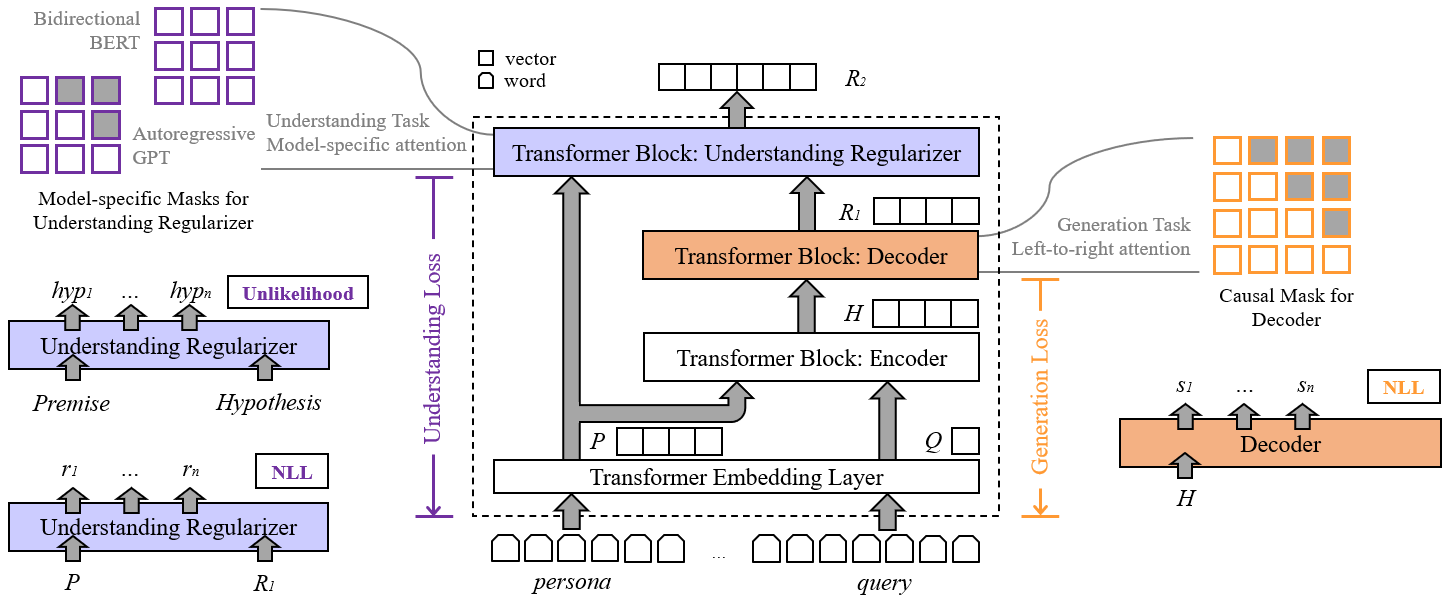}
\caption{An overview of the proposed stack-propagation framework \textit{EDU}. There are three Transformer blocks: an encoder (denoted as $\mathbb{E}$), a response generation decoder (denoted as $\mathbb{D}$) with causal attention mask, and a consistency understanding regularizer (denoted as $\mathbb{U}$) with model-specific attention mask, which depends on its initialization model. The tasks of personalized dialogue generation and consistency understanding are jointly modeled. Accordingly, two losses will be back-propagated, including a dialogue generation loss and a consistency understanding loss. The dialogue generation loss is negative log-likelihood (NLL) objective and back-propagates from $\mathbb{D}$ and $\mathbb{E}$ to the Transformer embedding layer. The consistency understanding loss is a combination of negative log-likelihood and unlikelihood objectives, where it back-propagates from the understanding regularizer $\mathbb{U}$ through $\mathbb{D}$ and $\mathbb{E}$ to the embedding layer. 
The architecture design and loss back-propagation strategy together form our stack-propagation framework.
}
\label{fig:3}
\end{figure}

\section{The Stack-Propagation Framework}
In this section, we present the stack-propagation EDU model, a novel Transformer-stacked framework that jointly models dialogue generation and consistency understanding for low-resource personalized dialogue generation. We first give an introduction about the settings of low-resource personalized dialogue generation (\cref{sec:task_overview}) and how we regularize dialogue generation with consistency understanding (\cref{sec:disentangle}). Then we present each sub-module in the framework (\cref{sec:framework}) as well as the training objectives (\cref{sec:training_objectives}). After that, we discuss different initialization schemes (\cref{sec:model_init}) of the EDU.

\subsection{Formulation of Low-resource Personalized Dialogue Generation}
\label{sec:task_overview}
In this work, we aim to address the limited personalized data  issue in persona-based dialogues. The restrictions on the availability of personalized data are reflected in: 1) using as little crowd-sourced personalized dialogue data as possible, and 2) not using the expensive dialogue-specific consistency understanding data.
Personalized dialogue data teaches the model how to generate dialogue responses and integrate persona information during training. The former is similar to the training of a language model, which can be learned from large-scale unlabeled texts rather than the personalized dialogue data itself. Compared with the language model training, the alignment between persona and response can be learned from much less personalized data. Therefore, the usage of crowd-sourced personalized dialogue data can be reduced.
On the other hand, to address the challenges of consistency understanding brought by limited data, we turn to leverage large-scale non-dialogue inference data by deliberately designing our stack-propagation framework.
We give a formal definition of the low-resource personalized dialogue generation task as follows:

\begin{table}[ht]
\centering
\caption{The notation table of this paper. We use different fonts to distinguish among different groups of notations.}
\begin{tabular}{@{}c|ll@{}}
\toprule
\textbf{Group} & \textbf{Notation} & \multicolumn{1}{c}{\textbf{Description}} \\ \midrule
\multirow{11}{*}{Model} & $\mathbb M$ & The proposed stack-propagation framework, i.e., EDU \\
 & $\mathbb E$ & The encoder in the EDU framework \\
 & $\mathbb{D}$ & The response generation decoder in the EDU framework \\
 & $\mathbb{U}$ & The understanding regularizer in the EDU framework, also a decoder \ \ \ \\
 & $NLL$ & The negative log-likelihood training objective \\
 & $UL$ & The unlikelihood training objective \\
 & $\alpha$ & A balance factor between positive and negative loss in $UL$ \\
 & $\beta$  & A balance factor between the losses from $\mathbb{D}$ and $\mathbb{U}$ \\
 & ${L}$ & The model training loss \\
 & $\theta$ & The learnable parameters in embedding layer, $\mathbb E$, and $\mathbb {D}$ \\
 & $\gamma$ & The learnable parameters in embedding layer, $\mathbb E$, $\mathbb {D}$, and $\mathbb {U}$ \\ \midrule
\multirow{12}{*}{Data} & $\mathcal P$ & The personas in personalized dialogue dataset \\
 & $\mathcal Q$ & The dialogue queries in personalized dialogue dataset \\
 & $\mathcal R$ & The dialogue responses in personalized dialogue dataset \\
 & $\mathcal{\hat R}$ & The generated responses by the model \\
 & $\mathcal N$ & The non-dialogue inference data, e.g., SNLI and MNLI \\
 & $\mathcal {\bar P}$ & The premise in the non-dialogue inference data \\
 & $\mathcal {\bar R}$ & The hypothesis in the non-dialogue inference data \\
 & $\mathcal N^+$ & The entailed <$\mathcal {\bar P}$, $\mathcal {\bar R}$> pairs in the non-dialogue inference data \\
 & $\mathcal N^-$ & The contradicted <$\mathcal {\bar P}$, $\mathcal {\bar R}$> pairs in the non-dialogue inference data \\
 & $q_1,q_2,...,q_n$ & A query in personalized dialogue dataset with $n$ words \\
 & $r_1,r_2,...,r_m$ & A response in personalized dialogue dataset with $m$ words \\
 & $\hat r_1,\hat r_2,...,\hat r_m$ & A generated dialogue response with $m$ words \\ \midrule
\multirow{6}{*}{Representation} & $P$ & The embeddings of personas \\
 & $Q$ & The embeddings of dialogue queries \\
 & $R$ & The embeddings of dialogue responses \\
 & $H$ & The hidden states of the encoder $\mathbb{E}$ \\
 & $R_1$ & The hidden states of the response generation decoder, i.e., $\mathbb{D}$ \\
 & $R_2$ & The hidden states of the consistency understanding regularizer, i.e., $\mathbb{U}$ \\ \bottomrule
\end{tabular}
\label{tab:notation}
\end{table}

We note the dialogue query as $\mathcal Q$ = $q_1,q_2,...,q_n$, the ground-truth response as $\mathcal R$ = $r_1,r_2,...,r_m$, and the personas as $\mathcal P$. In addition, let $\mathcal N$ denote the non-dialogue inference data, which consists of premise $\mathcal {\bar P}$, hypothesis $\mathcal {\bar R}$, and their entailment label. The label \textit{entailment} means the hypothesis can be inferred from the premise while \textit{contradiction} means cannot.
For simplicity and readability, in the following sections, we use fonts to distinguish the natural languages ($\mathcal P$, $\mathcal Q$, $\mathcal R$) from their vector representations ($P$, $Q$, $R_1$, $R_2$), as summarized in Table~\ref{tab:notation}.

The task of the low-resource personalized dialogue generation is to learn a dialogue model ${\mathbb M}$ to generate a persona-consistent response $\mathcal{\hat R}=\hat r_1,\hat r_2,...,\hat r_m$, conditioned on both persona $\mathcal P$ and dialogue query $\mathcal Q$, i.e., $\mathcal{\mathcal{\hat R}}={\mathbb M}(\mathcal P, \mathcal Q)$, where we try to minimize the usage of personalized dialogue data, and only non-dialogue inference data are available.
Figure~\ref{fig:3} shows an overview of the stack-propagation framework EDU. The proposed method $\mathbb M$ stacks three Transformer blocks, including an encoder {$\mathbb E$}, a response generation decoder $\mathbb{D}$, and a consistency understanding decoder $\mathbb{U}$.
Specifically, {$\mathbb E$} reads the embeddings of persona and query, i.e., $P$ and $Q$, and jointly encodes them into hidden states $H$. {$\mathbb{D}$} performs cross-attention on the encoder hidden states $H$ to model the context information, and generate a coarse response representation $R_1$. {$\mathbb{U}$} learns consistency understanding from non-dialogue inference data $\mathcal N$ and further converts $P$ and $R_1$ into final responses representations $R_2$. Finally, a persona-consistent response $\mathcal{\hat R}$ can be generated from $R_2$ through a linear layer with a softmax function, e.g., GPT, or a sequence of dot product over the embedding tables, e.g., BERT.

\subsection{Regularize Dialogue Generation with Consistency Understanding}
\label{sec:disentangle}
The biggest challenge brought by the lack of personalized data is that it is difficult to train dialogue generation and dialogue understanding models simultaneously.
Traditionally, a persona-based dialogue generation model requires persona $\mathcal P$ and dialogue query $\mathcal Q$ to generate a response. And a consistency understanding model requires persona $\mathcal P$, response $\mathcal R$, and the consistency labels between $\mathcal P$ and $\mathcal R$.
Moreover, the desired dataset for joint training should satisfy the format of \{$\mathcal P$, $\mathcal Q$, $\mathcal R$, Label\}, where both entailment-labeled and contradiction-labeled responses are needed.
Since the crowd-sourcing of personalized data and the annotation of consistency understanding data are already very expensive, it is not feasible to obtain sufficient annotated data that satisfy the ideal format of \{$\mathcal P$, $\mathcal Q$, $\mathcal R$, Label\}.

Instead of collecting an ideal dataset, we propose to disentangle dialogue generation and consistency understanding through regularizing generation with understanding.
In the proposed stack-propagation framework, we introduce the consistency understanding regularizer $\mathbb{U}$ to disentangle generation and understanding. $\mathbb{U}$ is also a decoder, but it reads persona embeddings $P$ and the hidden states $R_1$ from the first decoder, rather than the query embedding $Q$. Its outputs are the final response representation $R_2$.
The key to ``disentangling'' is $\mathbb{U}$ can get $R_1$ without the participation of $Q$, as $R_1$ is the representation of $\mathcal R$. Specifically, $\mathbb{U}$ uses the embedding of response, i.e., $R$, to approximate the hidden state from the first decoder, i.e., $R_1$. The main intuition behind such design is that we can restore the dialogue response from both $R$ and $R_1$, while we can get the embedding $R$ without the participation of dialogue query $Q$. As a result, the mapping from $R_1$ to $R_2$ in $\mathbb{U}$ can be independent of $Q$. 
In this way, it becomes possible to 1) learn persona-based dialogue generation from \{$\mathcal P$, $\mathcal Q$, $\mathcal R$\}, i.e., the personalized dialogue data, and 2) learn consistency understanding from \{$\mathcal P$, $\mathcal R$, Label\}. 
Moreover, to avoid the dependence on the limited consistency understanding data, we approximate \{$\mathcal P$, $\mathcal R$, Label\} by the abundant non-dialogue language inference data $\mathcal N$=\{\textit{Premise}, \textit{Hypothesis}, Label\}, where $\mathcal P$ and $\mathcal R$ are approximated by the \textit{Premise} and \textit{Hypothesis}, respectively. For $\mathbb{U}$, it can read the embeddings of  \textit{Premise} and \textit{Hypothesis} to reconstruct the representation of \textit{Hypothesis}, just like reading $P$ and $R_1$ to output $R_2$ in the dialogue generation process. 
Therefore, the introduction of consistency understanding regularizer $\mathbb{U}$ enables: 1) the usage of both personalized dialogue data and non-dialogue inference data under the same framework, and 2) the back-propagation of consistency supervisions to regularize the representation of responses.

Given persona $\mathcal P$ and response $\mathcal R$, suppose the regularizer $\mathbb{U}$ understands persona consistency, it should maximize the generation probability of $\mathcal R$ when $\mathcal R$ entails $\mathcal P$. Otherwise, it should minimize the generation probability of $\mathcal R$. 
Based on this intuition, we choose to apply unlikelihood objective~\cite{welleck2019neural} on $\mathbb{U}$ to make it understand consistency. In short, the unlikelihood objective would penalize an undesired sequence by minimizing their generation probabilities, and the detailed formulations of training objectives will be provided in~\cref{sec:training_objectives}.

 \subsection{EDU: Encoder, Decoder, and Understanding Regularizer}
\label{sec:framework}
As aforementioned, the proposed EDU framework employs an encoder $\mathbb E$ and two decoders, including a response generation decoder $\mathbb{D}$ and an understanding regularizer $\mathbb{U}$. During training, a negative log-likelihood loss for response generation will back-propagate from $\mathbb{D}$ through $\mathbb E$ to the embedding layer. For consistency understanding, an unlikelihood and reconstructive loss will back-propagate from $\mathbb{U}$ through $\mathbb{D}$ and $\mathbb E$ until the embedding layer. In this way, consistency understanding supervisions flow back to regularize the dialogue generation task and can be jointly learned with dialogue generation within the stack-propagation framework.
We will detail each sub-module as follows.

\subsubsection{Input Representation}
In the proposed framework, the input is a combination of persona $\mathcal P$ and dialogue query $\mathcal Q$. For the persona, we can always linearize it into a sequence of words, no matter $\mathcal P$ is personal facts (e.g., ``I have two dogs'') or profiles (e.g., ``location: Seattle''). A special token $[s]$ is added between the persona and dialogue query to provide segment information for the model, and the input sequence is formatted as:
\begin{gather}
\label{formula:input}
  input = p^{(0)}_1,p^{(0)}_2,...,p^{(t)}_{u_t},[s],q_1,q_2,...,q_n
\end{gather}
where $t$ is the number of persona texts, and $u_t$ is the number of words in the $t$-th persona text.
Then the embedding layer will convert $input$ into vector representations through tokenization and embedding lookup. We follow the usual practice to sum up the token embedding, type embedding, and position embedding as the final input embedding, where we set the type embedding to 0 for the persona and to 1 for the query. Note that $\mathcal P$ and $\mathcal Q$ can also get their representations independently. The resulted representations are $P$ and $Q$, and we jointly note them as $emb$ = $e^{p}_1,e^{p}_2,...,e^{q}_{l}$, where $l$ is the maximum length of the $input$.

\subsubsection{Backbone: Multi-Layer Transformer} Before going into each module in detail, we first brief our framework's backbone, i.e., the multi-layer Transformer, and unify the notation used in the following sections.
Transformer encodes input embeddings into contextual representations at different levels of abstract using a $N$-layer Transformer. In each Transformer layer, multiple self-attention heads are used to aggregate the output representations of the previous layer. For the $i$-th Transformer layer, the output of self-attention head $j$ $\textbf{A}^j_i$ is:
\begin{gather}
\label{formula:attention}
%   \textbf{Q}=\textbf{H}^{i-1}\textbf{W}_i^{Q}, \textbf{K}=\textbf{H}^{i-1}\textbf{W}_i^{K}, \textbf{V}=\textbf{H}^{i-1}\textbf{W}_i^{V} \\
  \textbf{A}^j_i = \mathrm{softmax}(\frac{\textbf{Q}\textbf{K}^\intercal}{\sqrt{d_k}}+\textbf{M})\textbf{V}
\end{gather}
where $\textbf{Q}$, $\textbf{K}$, and $\textbf{V}$ are query, key, and value in the $i$-th layer, respectively, $d_k$ is the dimension of the query and key, and $\textbf{M}$ is the mask matrix which determines whether a pair of tokens can be attended to each other by
\begin{gather}
\textbf{M}_{ij}=\left\{
\begin{aligned}
& 0,         & \text{allow to attend} \\
& -\infty,   & \text{prohibit from attending}
\end{aligned}
\right.
\end{gather}
Suppose the $i$-th token is not allowed to attend to the $j$-th token, the $\textbf{M}_{ij}$ will be ``$-\infty$'', and then according to equation~\ref{formula:attention} the attention score between the $i$-th and $j$-th tokens will approach zero. Thus, their attention is prevented. As shown in Figure~\ref{fig:3}, we apply both causal and bidirectional masks to control what context a token can attend to when computing its contextualized representation in different tasks.
Based on the attention function, Transformer further introduces a multi-head attention mechanism, which performs the attention function in parallel, yielding $d_v$-dimensional output values, where $d_v$ is the dimension of the value $\textbf{V}$. These values are concatenated and projected again to the final values. The multi-head attention in the $i$-th layer is
\begin{gather}
\label{formula:MultiHead}
  \mathrm{MultiHead}(\textbf{Q},\textbf{K},\textbf{V})=\mathrm{Concat}(\textbf{A}^1_i,...,\textbf{A}^h_i)W^O
\end{gather}
where $W^O$ is the projection parameter matrix and the dimension of $W^O$ is $hd_v \times d_{model}$, where $h$ is number of attention heads, $d_{model}$ is the dimension of the hidden states.
In addition to attention sub-layers, each of the Transformer layers contains a fully
connected feed-forward network, which is applied to each position separately and identically:
\begin{gather}
\label{formula:FFN}
  \mathrm{FFN}(x)=\mathrm{max}(0,xW_1+b1)W_2+b_2
\end{gather}
where two linear transformations with a ReLU activation in between are performed in the $\mathrm{FFN}$ function. The parameters in $\mathrm{FFN}$, i.e., $W_1$,$W_2$,$b_1$, and $b_2$, are different from layer to layer.

\subsubsection{Encoder}
The encoder $\mathbb E$ works like a standard Transformer encoder but formats the input tokens $\mathcal{P}$ and $\mathcal{Q}$ according to equation~\ref{formula:input}, and then the input embeddings are bidirectionally encoded to a sequence of hidden vectors, from which the downstream tasks will be performed on. In short:
\begin{gather}
\label{formula:encoder}
  \mathbb{E}(\mathcal{P},\mathcal{Q})={H}
\end{gather}
More specifically, we first get the input embeddings $emb$ of $\mathcal{P}$ and $\mathcal{Q}$. Then the encoder $\mathbb E$ will perform multi-head attention on the $emb$ to transform the embeddings into a sequence of hidden vectors $H$, followed up is a fully connected feed-forward network (FFN). There are $N$ identical layers in $\mathbb E$, and for $\mathbb E$'s hidden states $h^i$ in the $i$-th layer:
\begin{gather}
\label{formula:encoder_h}
  h^{i+1} = \mathrm{FFN}(\mathrm{MultiHead}(h^{i},h^{i},h^{i}))
\end{gather}
where $h^{1}$ = $emb$, and $h^{N}$ is the output of encoder.
The $\mathrm{MultiHead}$ and $\mathrm{FFN}$ are defined in equation~\ref{formula:MultiHead} and \ref{formula:FFN}, respectively.

\subsubsection{Response Generation Decoder}
The response generation decoder $\mathbb{D}$ is an auto-regressive Transformer decoder. First, a cross-attention is inserted between $\mathbb E$ and $\mathbb{D}$ to share the context representation $H$ encoded by $\mathbb E$. After that, a causal mask $M_{causal}$ is applied to $\mathbb{D}$ to preserve the auto-regressive generation property under different initialization schemes, which will be detailed in \cref{sec:model_init}. The output is the hidden state $R_1$, which can be used to decode response tokens with an output layer or embedding dot product. In short:
\begin{gather}
\label{formula:d1}
  \mathbb{D}(H,M_{causal})={R_1}
\end{gather}
The cross-attention interacts information between the Transformer encoder and decoder, and it is randomly initialized and updated during training.
In the cross-attention, the query is the outputs of the previous layer in $\mathbb{D}$, and the keys and values come from context information $H$. For $\mathbb{D}$'s hidden states $h_1^i$ in the $i$-th layer:
\begin{gather}
\label{formula:d1_multihead}
  h_1^{i+1} = \mathrm{FFN}(\mathrm{MultiHead}(h_1^{i}, H, H))
\end{gather}
Similar to the attention mechanism between encoder-decoder in the sequence-to-sequence model~\cite{sutskever2014sequence}, the cross-attention attends to all positions in the context representations $H$ to allow $h_1$ aggregate context information. 
In training, $h_1^{1}$ is initialized from the embeddings of the ground-truth response:
\begin{gather}
\label{formula:d1_multihead_2}
  h_1^{1} = \mathrm{FFN}(\mathrm{MultiHead}(R, R, R))
\end{gather}
where $R$ is the embedding of the target response with a start token. Future tokens in the target response should not be leveraged at each generation step in both training and inference. Therefore, as shown in Figure~\ref{fig:3}, a causal mask is applied to $\mathbb{D}$ to ensure that the token generation is auto-regressive. 
$\mathbb{D}$ also has $N$ identical layers. And $R_1$ is the hidden state of the last layer $h_1^{N}$, which will be further fed to the consistency understanding decoder $\mathbb{U}$.

\subsubsection{Consistency Understanding Regularizer}
Similar to $\mathbb{D}$, the consistency understanding regularizer $\mathbb{U}$ is also a $N$-layer Transformer decoder. Regularizer $\mathbb{U}$ takes persona embedding $P$ and the dialogue response representation's approximation $R^\ast$ as input, and $\mathbb{U}$'s output is the final representation for response generation $R_2$. Since both the masked language model and auto-regressive language model are capable of language understanding, we make no assumption about the shape of the attention masks in the EDU framework. According to the initialization language model type, we apply the model-specific mask to the understanding regularizer. In short:
\begin{gather}
\label{formula:d2}
  \mathbb{U}(P, R^\ast,M_{model-specific})={R_2}
\end{gather}
where $R^\ast$=$R_1$ when trained with personalized dialogue data, and $R^\ast$=$R$ when trained with non-dialogue inference data. 
% As aforementioned (\cref{sec:disentangle}), when trained with inference data the $R$ is from the shared embedding layers and thus does not need to flow through $\mathbb{E}$ and $\mathbb{D}_1$.
For the $i$-th layer in $\mathbb{U}$, the multi-head attention is performed twice to update its hidden state $h_2^i$:
\begin{gather}
\label{formula:d2_multihead}
  p^{i+1} = \mathrm{FFN}(\mathrm{MultiHead}(h_2^{i}, P, P))\\
  h_2^{i+1}=\left\{
  \begin{aligned}
    & \mathrm{FFN}(\mathrm{MultiHead}(p^{i+1}, R_1, R_1))         & \text{for personalized dialogue data} \\
    & \mathrm{FFN}(\mathrm{MultiHead}(p^{i+1}, R, R))   & \text{for non-dialogue inference data}
  \end{aligned}
  \right.
\end{gather}
For the personalized dialogue data, they are in a \{\textit{persona}, \textit{query}, \textit{response}\} format, and these data can flow through $\mathbb{E}$, $\mathbb{D}$ to the $\mathbb{U}$. Therefore, here $R_1$ is the hidden states of $\mathbb{D}$. While for the non-dialogue inference data, they only have premises and hypotheses, where the premise and hypothesis approximate persona and response, respectively. As there is no query in the inference data, we get the representations of the approximated response through its embeddings, i.e., $R$.
The resulted $h_2^{i+1}$ in each layer thus fuses information from both persona and response. The output of $\mathbb{U}$'s $N$-th layer is the final representation $R_2$. 
Once we get $R_2$, we can 1) get the generated response ${\mathcal {\hat R}}$ through a linear layer with softmax or embedding dot product, and 2) penalize the approximated responses labeled as a contradiction in the inference data to obtain a consistency understanding capability.

\subsection{Training Objectives}
\label{sec:training_objectives}
One of the key features that distinguish the stack-propagation from pipeline stacking is the loss back-propagation strategy, as discussed in \cref{sec:stack_propagation}. 
The proposed stack-propagation framework jointly models dialogue generation and consistency understanding tasks and treats consistency understanding as a regularizer of dialogue generation.
To this end, the consistency understanding loss first flows through the understanding regularizer $\mathbb{U}$, and is further back-propagated to the decoder $\mathbb{D}$ and encoder $\mathbb E$, until the embedding layer. For the dialogue generation loss, it is back-propagated from $\mathbb{D}$ to $\mathbb{E}$ and embedding layer, as the usual practice in a sequence-to-sequence model.

Specifically, we employ the negative log-likelihood (NLL) objective for dialogue generation and the unlikelihood (UL) objective for consistency understanding. A brief illustration of which module the objective is applied to is shown in Figure~\ref{fig:3}, and detailed descriptions will be provided in the following section.

\paragraph{Negative Log-Likelihood for Response Generation}
To generate dialogue responses, a dialogue model will maximize the probabilities of every target token conditioned on the hidden states, i.e., $R_1$ or $R_2$, produced by the model.
In our framework, we apply the widely used negative log-likelihood loss for response generation training. 
The encoder $\mathbb E$ reads the persona $\mathcal P$ and query $\mathcal Q$ and pass $H$ into $\mathbb{D}$ to predict the target response $\mathcal R$ from the coarse representations $R_1$:
\begin{equation}
\begin{aligned}
\label{formula:nll_d}
  {L}^{\mathbb{D}}_{NLL} &= -\mathrm{log}(p_\theta(\mathcal R|\mathcal P,\mathcal Q)) = -\sum_{i=1}^{|\mathcal R|}\mathrm{log}(p_\theta(r_i|\mathcal P,\mathcal Q,\mathcal R_{<i}))
\end{aligned}
\end{equation}
where the $\theta$ is the learnable parameters in $\mathbb{D}$, $\mathbb E$ and the embedding layer.
The generation part in $\mathbb{U}$ is also trained by the negative log-likelihood objective in a manner like auto-encoding. The regularizer $\mathbb{U}$ reads persona embeddings $P$ and raw representations $R_1$ to predict the target response $\mathcal R$:
\begin{equation}
\begin{aligned}
\label{formula:nll_u}
  {L}^{\mathbb{U}}_{NLL} &= -\mathrm{log}(p_\gamma(\mathcal R|P, R_1)) = -\sum_{i=1}^{|\mathcal R|}\mathrm{log}(p_\gamma(r_i|P,R_1,\mathcal R_{<i}))
\end{aligned}
\end{equation}
where the $\gamma$ is the learnable parameters in $\mathbb{E}$, $\mathbb{D}$, $\mathbb U$, and the embedding layer. As $\theta$ is a subset of $\gamma$, i.e., $\theta \in \gamma$, the gradients from the understanding regularizer $\mathbb{U}$ will also flow back to the dialogue generation sub-modules.
% making the proposed EDU architecture a stack-propagation framework.

\paragraph{Unlikelihood Objective for Consistency Understanding}
As an important step to alleviate the dependence on dialogue-specific consistency understanding data, we turn to leverage the abundant non-dialogue inference datasets.
Practically, we collect positive data $\mathcal N^+$ from the \textit{entailment} category in the non-dialogue inference dataset and collect negative data $\mathcal N^-$ from the \textit{contradiction} category:
\begin{gather}
\label{formula:8}
  \mathcal N^+ = \{(\mathcal{\bar P}^{(i)},\mathcal{\bar R}^{(i)+})\},\ \ \mathcal N^- = \{(\mathcal{\bar P}^{(j)},\mathcal{\bar R}^{(j)-})\}
\end{gather}
where $\mathcal{\bar P}$ and $\mathcal{\bar R}$ are premise and hypothesis from the non-dialogue inference data, respectively. Here we leverage the premise $\mathcal{\bar P}$ to approximate the persona $\mathcal{P}$ and the hypothesis $\mathcal{\bar R}$ to approximate the response $\mathcal{R}$. Due to regularizer $\mathbb{U}$ disentangles the representations of query and response, the inference data can still be used to train the model even if there is no query in its format.
The representations of $\mathcal{\bar P}$ and $\mathcal{\bar R}$ in the EDU framework are denoted as $\bar P$ and $\bar R$, which can be obtained through the embedding layer.
The key idea behind unlikelihood training is decreasing the model's probability of undesired tokens.
For data from $\mathcal N^+$, we still apply the negative log-likelihood loss:
\begin{equation}
\begin{aligned}
\label{formula:9}
  {L}^{\mathbb{U}^+}_{UL} = -\sum_{i=1}^{|\mathcal {\bar R}|}\mathrm{log}(p_\gamma(\bar r_i|\bar P,\bar R,\mathcal {\bar R}_{<i}))
\end{aligned}
\end{equation}

For the training data from $\mathcal N^-$, we apply an unlikelihood loss to minimize the likelihood of contradictions by reversing the generation probabilities:
\begin{equation}
\begin{aligned}
\label{formula:ul}
  {L}^{\mathbb{U}^-}_{UL} = -\sum_{i=1}^{|\mathcal {\bar R}|}\mathrm{log}(1-p_\gamma(\bar r_i|\bar P,\bar R,\mathcal {\bar R}_{<i}))
\end{aligned}
\end{equation}
which penalizes every token in the contradicted target. Therefore, the loss ${L}^{\mathbb{U}^-}_{UL}$ makes generating contradicted responses less likely.
The final unlikelihood training objective is a weighted sum of the two parts of losses:
\begin{equation}
\begin{aligned}
\label{formula:unlikelihood_sum}
  {L}_{UL} = {L}^{\mathbb{U}^+}_{UL}+\alpha {L}^{\mathbb{U}^-}_{UL} = -\sum_{i=1}^{|\mathcal {\bar R}|}\mathrm{log}(p_\gamma(\bar r_i|\bar P,\bar R,\mathcal {\bar R}_{<i})) -\alpha\sum_{i=1}^{|\mathcal {\bar R}|}\mathrm{log}(1-p_\gamma(\bar r_i|\bar P,\bar R,\mathcal {\bar R}_{<i}))
\end{aligned}
\end{equation}
where $\alpha$ is a balance factor that controls the information from positive and negative examples.

\paragraph{Training Procedure}
After we have introduced the framework architecture and training objectives, we summarize the training procedure as follows:
\begin{enumerate}
\item [1)] Response Generation. Given $\mathcal P$, $\mathcal Q$, and $\mathcal R$ from personalized dialogue data, we calculate the negative log-likelihood loss of response generation ${L}_{1}={L}^{\mathbb{D}}_{NLL}+\beta {L}^{\mathbb{U}}_{NLL}$ according to equation~\ref{formula:nll_d} and \ref{formula:nll_u}, where $\beta$ is a hyperparameter and is usually set to less than 1;
\item [2)] Consistency Understanding. Given $\mathcal N^+$ and $\mathcal N^-$ from non-dialogue inference data, we calculate the unlikelihood loss for consistency understanding ${L}_{2}={L}_{UL}$ according to equation~\ref{formula:unlikelihood_sum};
\item [3)] Optimization. Sum up ${L}_{1}$ and ${L}_{2}$, and update parameters according to the back-propagated gradients.
\end{enumerate}
To summarize, the training loss is:
\begin{equation}
\begin{aligned}
\label{formula:training_loss}
  {L}   &= L_1 + L_2 
        = {L}^{\mathbb{D}}_{NLL}+\beta {L}^{\mathbb{U}}_{NLL} + L_{UL}
        = \left({L}^{\mathbb{D}}_{NLL} + \beta {L}^{\mathbb{U}}_{NLL}\right) + {\left({L}^{\mathbb{U}^+}_{UL} + \alpha {L}^{\mathbb{U}^-}_{UL}\right)}
\end{aligned}
\end{equation}
where the first group of is the negative log-likelihood loss from equation~\ref{formula:nll_d} and \ref{formula:nll_u}, and the second group is the unlikelihood loss from equation~\ref{formula:9} and \ref{formula:ul}.

\paragraph{Implementation Details}
We report the configurations of the used pre-trained language models in Table~\ref{tab:model_config} and the optimization hyperparameters of training in Table~\ref{tab:hyperparam}.
Empirically, we set the maximum length of persona, dialogue query, and response in a single turn to 64, 64, and 32, respectively. We set $\alpha$ to 1e-2 and $\beta$ to 0.8.
In the main experiments, for a fair comparison to the baselines, all the Transformer blocks are initialized from the 12 layers 12 attention-heads BERT$_\text{base}$ and GPT models, with a hidden size of 768 and feed-forward size of 3,072. Moreover, we further explore the potential of the framework through initializing with stronger and larger language models, including RoBERTa$_\text{base}$, BERT$_\text{large}$, and GPT-2$_\text{medium}$. The framework's training with the 110M and 340M language models requires about 20G (batch size 32) and 70G (batch size 64) GPU memories on a single GPU, respectively. Here we use Tesla V100 32G and A100 80G GPUs for the framework's training.
We implement the EDU framework with PyTorch\footnote{https://pytorch.org/} and HuggingFace's transformers library\footnote{https://github.com/huggingface/transformers}.
Before training, since the cross-attention between the encoder and decoder does not exist, we first warm up the encoder and decoder with the loss in equation~\ref{formula:nll_d} by three epochs.
For the hyperparameter selection, we run a grid search on the \textit{batch size} and \textit{learning rate}, where the batch size is from [32, 64] and the learning rate is from [1e-5, 2e-5, 3e-5, 4e-5, 5e-5]. 
We run the training on each combination for up to 30 epochs according to the overall loss in equation~\ref{formula:training_loss} and select the model with the best automatic metrics on the development dataset.

\begin{minipage}[t]{0.65\textwidth}
\small
\centering
\captionsetup{type=table}
\caption{Configuration of the pre-trained language models for initializing. We also explore the framework's potential by initializing from larger size models.}
\begin{tabular}{lccccc}
\toprule
\textbf{Config.} & \textbf{BERT$_\text{base}$} & \textbf{GPT} & \textbf{RoBERTa} & \textbf{BERT$_\text{large}$} & \textbf{GPT2$_\text{medium}$} \\ \midrule
Layers & 12 & 12 & 12 & 24 & 24 \\
Hidden size & 768 & 768 & 768 & 1024 & 1024 \\
Parameters & 110M & 110M & 125M & 340M & 345M \\
Heads & 12 & 12 & 12 & 16 & 16 \\
Dropout & 0.1 & 0.2 & 0.1 & 0.1 & 0.2 \\
GPUs & V100x1 & V100x1 & V100x1 & A100x1 & A100x1 \\
Training Mem. & 20G & 20G & 20G & 70G & 70G \\
\bottomrule
\end{tabular}
\label{tab:model_config}
\end{minipage}
\hfill
\begin{minipage}[t]{0.3\textwidth}
\small
\centering
\captionsetup{type=table}
\caption{Optimization hyperparameters. ``/'' denotes base/large models' values.}
\begin{tabular}{lr}
\toprule
\textbf{Hyperparameters} & \textbf{Value} \\ \midrule
Batch size & 32 / 64 \\
Optimizer & Adam \\
Learning rate & 5e-5 / 1e-5 \\
Learning Rate Decay & Linear \\
Adam $\epsilon$ & 1e-8 \\
Adam $\beta$ & (0.9, 0.999) \\
Weight decay & 0.01 \\
\bottomrule
\end{tabular}
\label{tab:hyperparam}
\end{minipage}

\subsection{Framework Initialization Schemes}
\label{sec:model_init}
There are two approaches to initialize of the proposed EDU framework: 1) randomly initialize all parameters and train the model on personalized dialogue data from scratch, and 2) initialize the model parameters from pre-trained language models and then finetune on the personalized dialogue data.
In the second approach, typically we can leverage auto-regressive language models, e.g., GPT, and masked language models, e.g., BERT. As discussed in \cref{sec:auto_and_maksed}, the auto-regressive language models are naturally compatible with generation tasks, while the masked language models are better at understanding tasks. Since the proposed stack-propagation framework regularizes dialogue generation with consistency understanding, we initialize the framework with different types of language models to thoroughly investigate the framework's potential. Specifically, according to the initialization approaches of encoder ${\mathbb E}$, decoder ${\mathbb{D}}$, and regularizer ${\mathbb{U}}$, we explore the following framework initialization schemes:
\begin{enumerate}
\item [1)] BoB (BERT-over-BERT). It initializes ${\mathbb E}$, ${\mathbb{D}}$, and ${\mathbb{U}}$ with the same BERT checkpoint and randomly initializes the cross-attention between ${\mathbb E}$ and ${\mathbb{D}}$. As BERT is a typical masked language model, we apply a causal mask to the ${\mathbb{D}}$ to keep the auto-regressive generation manner. We note the model initialized under this scheme as \textbf{EDU$_{\text{BoB}}$}.
\item [2)] GoG (GPT-over-GPT). Similar to the BoB scheme, it initializes ${\mathbb E}$, ${\mathbb{D}}$, and ${\mathbb{U}}$ with the same openAI GPT~\cite{radford2018improving} checkpoint, and the cross-attention between ${\mathbb E}$ and ${\mathbb{D}}$ is randomly initialized. Since GPT is an auto-regressive model, it only models the left-to-right information when applied for language understanding tasks. We will see the performance differences brought by the unidirectional information in the following experiments. And we note the model initialized under this scheme as \textbf{EDU$_{\text{GoG}}$}. 
\item [3)] BGB (BERT-GPT-BERT). This scheme initializes different sub-modules with different language models according to the module's characteristics. For ${\mathbb E}$ and ${\mathbb{U}}$, both of them need a bidirectional representation to capture full semantics, and thus we leverage the BERT model to initialize them. For ${\mathbb{D}}$, it needs an auto-regressive manner to generate responses, and thus we initialize it with the GPT model. The cross-attention between ${\mathbb E}$ and ${\mathbb{D}}$ is also randomly initialized as such attention does not exist in BERT and GPT. We note this scheme as \textbf{EDU$_{\text{BGB}}$}.
\item [4)] ToT (Transformer-over-Transformer). It randomly initializes all parameters in the three Transformer blocks (details in \cref{sec:framework}) and trains them on the personalized data from scratch. We set this initialization scheme to investigate the role of language model pre-training in the stack-propagation framework. A BERT-like bidirectional mask is applied to the regularizer ${\mathbb{U}}$ in this scheme. We note this baseline scheme as \textbf{EDU$_{\text{ToT}}$}.
\end{enumerate}

\section{Experiments}
In this section, we conduct experiments to evaluate the effectiveness of the proposed EDU framework on two low-resource scenarios: 1) a persona-dense scenario, where all conversations are grounded on specific personas by crowd-sourcing data collection. We gradually reduce the number of examples by halving the training dataset to simulate a low-resource scenario. And 2) a persona-sparse scenario, where every conversation is associated with a specific profile, but only very few conversations are related to the profile. Persona-sparse is a typical scenario for the large-scale dialogue datasets collected from social media. In the following section, we first introduce the experimented datasets for two scenarios (\cref{sec:exp_datasets}). Then we brief the compared methods (\cref{sec:exp_baseline}), and next we discuss the evaluation metrics for both dialogue quality and persona consistency (\cref{sec:exp_metrics}). After that, we present the evaluation results of the persona-dense scenario (\cref{sec:exp_dense_results}), and we conduct analyses to have a more thorough view of the framework's performance. We further present validation experiments under the persona-sparse scenario (\cref{sec:exp_sparse_results}). At last, we showcase some examples to intuitively demonstrate how the EDU framework works (\cref{sec:exp_cases}).

\begin{table}[t]
\centering
\caption{Summaries of persona-based dialogue datasets in the experiments. ``\#'' counts the number of query-response pairs.}
\begin{tabular}{llllllll}
\toprule
\multicolumn{1}{l}{\textbf{Dataset}} &
  \multicolumn{1}{l}{\textbf{Type}} &
  \multicolumn{1}{l}{\textbf{Persona}} &
  \multicolumn{1}{l}{\textbf{From}} &
  \multicolumn{1}{l}{\textbf{Language}} &
  \multicolumn{1}{l}{\textbf{\# Train}} &
  \multicolumn{1}{l}{\textbf{\# Valid}} &
  \multicolumn{1}{l}{\textbf{\# Test}} \\ \midrule
PersonaChat      & Dense         & Personal Facts  & Crowd-Sourcing   & English           & 121,880           & 9,558             & 7,801            \\
PersonalDialog    & Sparse        & Profiles   & Social Media      & Chinese           & 5,014,349         & 423,817           & 10,000 / 521     \\ \bottomrule
\end{tabular}
\label{tab:dialogue_datasets}
\end{table}

\begin{table}[]
\centering
\caption{The splitting details of PersonaChat under the low-resource settings. Q-R denotes Query-Response.}
\label{tab:low_resource}
\resizebox{.835\textwidth}{!}{%
\begin{tabular}{llllll}
\toprule
\multicolumn{1}{l}{\textbf{Data Splitting}} & 
\multicolumn{1}{l}{\textbf{Dialogue IDs}} & 
\multicolumn{1}{l}{\textbf{\# Personas}} &
\multicolumn{1}{l}{\textbf{\# Dialogues}} &
\multicolumn{1}{l}{\textbf{\# Q-R Pairs}} &
\multicolumn{1}{l}{\textbf{\# Test Pairs}} \\ \midrule
Full Data & [1, 17,878] & 80,365 & 17,878 & 131,438 & 7,801 \\
1/2 Data  & [1, 8,939]  & 40,245 & 8,939 & 65,719 & 7,801 \\
1/4 Data  & [1, 4,459]  & 20,070 & 4,459 & 32,859 & 7,801 \\
1/8 Data  & [1, 2,226]  & 10,027 & 2,226 & 16,430 & 7,801 \\ \bottomrule
\end{tabular}
}
\end{table}

\subsection{Datasets}
\label{sec:exp_datasets}
We evaluate the performance of the stack-propagation framework on a persona-dense scenario and a persona-sparse scenario with two publicly available personalized datasets:
\begin{itemize}
  \item {\bf PersonaChat}~\cite{zhang-etal-2018-personalizing} is a well-known personalized dataset. It is collected from crowd-sourced workers and thus covers rich persona information. Each dialogue in this dataset is grounded on certain personas. We use the ConvAI2 PersonaChat~\cite{dinan2019second}, so the results are comparable to previous methods. Besides the full dataset experiments, we reduce the number of training examples to 1/2, 1/4, and 1/8 scales to simulate a low-resource setting.
  \item {\bf PersonalDialog}~\cite{zheng2019personalized} is a large-scale persona-sparse dataset. It is collected from the popular Chinese social media Weibo. Every dialogue in the PersonalDialog is associated with users' profiles, i.e., persona, but the majority are not grounded on the persona. There are two testsets: 1) a random testset, which is identically distributed as the persona-sparse training data. 2) a biased testset, which is manually filtered to be persona-relevant.
\end{itemize}
We report the summary of the two persona-based dialogue datasets in Table~\ref{tab:dialogue_datasets} and present the details of the low-resource setting of PersonaChat in Table~\ref{tab:low_resource}.

\begin{table}[t]
\centering
\caption{Summaries of different inference datasets. ``\#'' counts the number of <premise, hypothesis, label>.}
\begin{tabular}{lllllll}
\toprule
\multicolumn{1}{l}{\textbf{Dataset}} &
  \multicolumn{1}{l}{\textbf{Type}} &
  \multicolumn{1}{l}{\textbf{Usage}} &
  \multicolumn{1}{l}{\textbf{Language}} &
  \multicolumn{1}{l}{\textbf{\# Entailed}} &
  \multicolumn{1}{l}{\textbf{\# Neutral}} &
  \multicolumn{1}{l}{\textbf{\# Contrad.}} \\ \midrule
MNLI  & Non-Dialogue & Training & English & 130,615 & 130,590 & 130,590 \\
CMNLI & Non-Dialogue & Training & Chinese & 130,612 & 130,555 & 130,616 \\ \midrule
DNLI  & Dialogue & Evaluation & English & 15,495  & 20,927  & 16,488  \\
KvPI  & Dialogue & Evaluation & Chinese & 33,114  & 54,426  & 31,000  \\
DECODE  & Dialogue & Evaluation & English & -  & - & 35,426  \\ \bottomrule
\end{tabular}
\label{tab:inference_datasets}
\end{table}

To tackle the consistency understanding issue brought by limited personalized data, we leverage large-scale non-dialogue inference data. We use the natural language inference dataset MNLI~\cite{williams-etal-2018-mnli} for the PersonaChat experiments and the Chinese CMNLI~\cite{xu-etal-2020-clue} for the PersonalDialog experiments. 
Moreover, to automatically compare the persona consistency performance of different models, we also leverage dialogue inference datasets, DNLI~\cite{WelleckDNLI}, KvPI~\cite{song-etal-2020-profile}, and DECODE~\cite{nie2021dolphin}, for evaluations.
We report the summaries of these inference datasets in Table~\ref{tab:inference_datasets}.
For the DNLI dataset, we only count the tuples that can be restored as the evaluation format of \{\textit{persona}, \textit{query}, \textit{response}, \textit{label}\} in our experiments.
For the DECODE dataset, we use the \textit{aggregated contradiction} instances as negative examples. 
% We sample dialogues other than the contradiction as non-contradiction ones, which does not distinguish between entailment and neutral.

\subsection{Compared Methods}
\label{sec:exp_baseline}
We compare the proposed stack-propagation framework with delicately designed non-pretrained models and strong pre-trained dialogue models.
We mainly compare the stack-propagation framework with the following models:
\begin{itemize}
    \item{\textbf{Transformer.}} The encoder-decoder-based Transformer~\cite{vaswani2017attention} is employed as a non-pretrained baseline for both experiments. We concat personas and dialogue queries as the model's inputs.

    \item{\textbf{CMAML.}} Meta-learning has been proven to be an effective way for few-shot learning. {CMAML}~\cite{song-etal-2020-learning} is a meta-learning-based method that learns to customize model structures to adapt from few-shot personas.
    \item{\textbf{GDR.}} Short for generate-delete-rewrite, {GDR}~\cite{song-etal-2020-generate} is a multi-stage generation method for personalized dialogues, which aims at improving persona consistency. The GDR model, along with the CMAML, is delicately designed for the persona-dense dataset, and thus we only compare with them on the PersonaChat dataset.
    \item{\textbf{LIC.}} Short for lost in conversation, {LIC}~\cite{golovanov2020lost} is the best performing model in the ConvAI2 PersonaChat challenge~\cite{dinan2019second}. LIC is a  pre-trained dialogue model with a general language model architecture, and thus we compare this model on both PersonaChat and PersonalDialog datasets. For more information about the models in ConvAI2 please refer to the competition technical report~\cite{dinan2019second}.
    \item{\textbf{AttentionRouting.}} AttentionRouting~\cite{Zheng_Zhang_Huang_Mao_2020} is a pre-trained model specially designed for the persona-sparse scenario, and it is also the latest model on the PersonalDialog dataset. 
    \item{\textbf{GPT-2.}} We also finetune a pre-trained {GPT-2} model~\cite{radford2019language} for a thorough comparison on PersonaChat.
\end{itemize}
Besides conventional methods, in section~\ref{sec:dense:further-explore}, we also compare the framework with fine-tuned  \textbf{BART}~\citep{lewis2020bart} and \textbf{T5}~\citep{T5}.

\subsection{Evaluation Metrics}
\label{sec:exp_metrics}
The goal of personalized dialogues is to generate natural, informative, and persona-consistent dialogue responses. 
According to such characteristics, we focus on the evaluations of two main aspects in the generated responses: {\it response quality} and {\it persona consistency}. To comprehensively compare different models, we employ both objective automatic metrics and subjective human evaluations.

\paragraph{Automatic Metrics} For dialogue quality, we employ perplexity ({\bf PPL}) and distinct 1/2 ({\bf Dist.1/2}) following common practice~\cite{zhang-etal-2018-personalizing,Zheng_Zhang_Huang_Mao_2020}.
Perplexity is defined as the exponential average negative log-likelihood of a sequence. Given the target response $\mathcal R$, and the corresponding persona $\mathcal P$ and query $\mathcal Q$, the perplexity of $\mathcal R$ is:
\begin{equation}
\begin{aligned}
\label{formula:ppl}
  \mathrm{PPL}\left(\mathcal{R}\right) = \mathrm{PPL}\left(\mathcal{R}|\mathcal{P},\mathcal{Q}\right) = \mathrm{exp}\left\{-\frac{1}{|\mathcal R|}\sum_{i=1}^{|\mathcal R|}\mathrm{log}(p(r_i|\mathcal P,\mathcal Q,\mathcal R_{<i}))\right\}
\end{aligned}
\end{equation}
Lower perplexity indicates better language modeling, and we report the averaged perplexity over every target response in the testset. Besides perplexity, distinct 1/2~\cite{li2016diversity} calculate the ratio of distinct uni-grams/bi-grams, respectively. A higher distinct value means more unique words and phrases, indicating a better diversity in the generated responses.

For persona consistency, we employ two metrics. The first one is Consistency Score ({\bf C.Score})~\cite{madotto-etal-2019-personalizing}, which leverages a referee model to predict consistency relations between personas and responses, and then a quantitative score is calculated according to the predicted relations. The C.Score can be defined as:
\begin{align}
\label{formula:11}
\begin{aligned}
\mathrm{NLI}(r,p_i)&=\left\{
\begin{aligned}
-1 & , & \text{if}\ r\ \text{contradicts}\ p_i \\
0 & , & \text{if}\ r\ \text{is irrelevant to}\ p_i \\
1 & , & \text{if}\ r\ \text{entails}\ p_i
\end{aligned}
\right.\\
\text{C.Score}(r) &= \sum\nolimits_{i=1}^t\mathrm{NLI}(r,p_i)
\end{aligned}
\end{align}
where $r$ is the generated response, and $t$ is the number of personas associated with each dialogue, and the NLI is the referee model.
Here we apply a pre-trained RoBERTa~\cite{liu2019roberta}, which is finetuned with the dialogue inference datasets, i.e., DNLI and KvPI, as the referee model. For accuracy, the referee model achieves 89.3\% and 88.9\% on DNLI and KvPI, respectively. The results align with the previously reported 88.20\%~\cite{WelleckDNLI} and 88.0\%~\cite{song-etal-2020-profile}.
The calculation of the C.Score is based on the response and its corresponding persona, where a persona usually consists of 3 to 5 persona texts.
We report the averaged C.Score over all generated responses and their personas in the testset.

The second metric for consistency evaluation is Delta Perplexity ({$\Delta$}{\bf P}), which evaluates consistency from the model's internal distributions.
For the entailed and contradicted dialogues in the dialogue inference dataset, we can calculate two kinds of perplexities: perplexity of entailed ({\bf p.Ent}) and contradicted ({\bf p.Ctd}).
Since perplexity is related to the generation probability, a dialogue model with good understanding capability should assign lower perplexity to the entailed dialogues while higher perplexity to the contradictions. From this intuition, the {$\Delta$}{P} can be defined as:
\begin{equation}
\label{formula:12}
\begin{aligned}
{\Delta \text P} = \text{p.Ctd} - \text{p.Ent} = \text{PPL}(\text{Contradicted}) - \text{PPL}(\text{Entailed})
\end{aligned}
\end{equation}
where \textit{Entailed} stands for the responses from the entailed \{\textit{persona}, \textit{query}, \textit{response}\} tuples, and so does the {\it Contradicted}, and the PPL is defined in equation~\ref{formula:ppl}. 
% In our experiments, we get the entailed and contradicted \{\textit{persona}, \textit{query}, \textit{response}\} tuples from the dialogue inference datasets DNLI and KvPI.
A larger {$\Delta$}{P} means the model can better distinguish entailment from contradiction. We report the averaged p.Ent and p.Ctd over all entailed and contradicted tuples from the dialogue inference datasets. And {$\Delta$}{P} is the difference between the averaged p.Ent and p.Ctd.

\paragraph{Human Evaluations}
We recruit two annotation teams from a third-party company; one team is for the English PersonaChat, and another is for the Chinese PersonalDialog. Each team consists of five professional annotators. These annotators have high-level language skills and well understand the annotation criteria, but they have no information about the models. 
These annotators are instructed to evaluate dialogue quality from three criteria, including fluency ({\bf Flue.}), informativeness ({\bf Info.}), and relevance ({\bf Relv.}). Each criterion is rated on a five-scale, where \textbf{1} means unacceptable, \textbf{3} means moderate, and \textbf{5} means perfect performance. \textbf{2} and \textbf{4} can be used for uncertain cases.
The three criteria are evaluated independently. For example, for the dialogue query ``how old are you?'', the response ``I'm a stay at home mom with three kids.'' should get 5 on both fluency and informativeness but be rated 1 on its relevance.
The annotators are also instructed to evaluate the consistency ({\bf Per.C.}) between persona and response, where \textbf{+1} means persona-relevant and consistent, \textbf{0} means persona-irrelevant, and \textbf{-1} means contradicted to the persona.
We randomly sample 100 \{\textit{persona}, \textit{query}, \textit{response}\} tuples from each model's generation results for evaluation under every setting.

\subsection{Evaluation Results for the Persona-Dense Scenario}
\label{sec:exp_dense_results}
In this section, we evaluate the EDU framework's performance under the persona-dense scenario with the PersonaChat dataset. We first report the full evaluation results (\cref{sec:dense-full}). Then we report the performance under a low-resource setting by gradually halving the personalized training examples (\cref{sec:dense-low-resource}). Finally, we conduct several analyses on the proposed framework (\cref{sec:dense-analysis}) and make further explorations to discuss the framework's potential (\cref{sec:dense:further-explore}).

\begin{table}[t]
\centering
\captionsetup{width=.743\textwidth}
\caption{Automatic evaluation results on the PersonaChat dataset, including response quality (left) and persona consistency (right). D.AVG is the average of Dist.1 and 2. The best results are in bold.}
\label{tab:dense_automatic}
\begin{tabular}{@{}l|rrrr|rrrr@{}}
\toprule
\textbf{Model} & \textbf{PPL} & \textbf{Dist.1} & \textbf{Dist.2} & \textbf{D.AVG} & \textbf{p.Ent} & \textbf{p.Ctd} & \textbf{$\Delta$P} & \textbf{C.Score} \\ \midrule
Transformer~\cite{vaswani2017attention} & 28.8 & 3.14 & 17.80 & 10.47 & 31.5 & 35.5 & 4.0 & 1.20 \\
CMAML~\cite{song-etal-2020-learning} & 36.7 & 1.00 & 2.10 & 1.55 & 32.3 & 37.5 & 5.2 & 6.96 \\
GDR~\cite{song-etal-2020-generate} & 16.7 & 3.76 & 23.10 & 13.43 & 19.7 & 32.3 & 12.6 & 7.89 \\
LIC~\cite{golovanov2020lost} & 17.3 & 6.29 & 28.99 & 17.64 & 13.7 & 20.4 & 6.7 & 14.12 \\
GPT-2~\cite{radford2019language} & 14.4 & 7.29 & 28.12 & 17.71 & 12.0 & 20.2 & 8.2 & 15.88 \\ \midrule
EDU$_{\text{ToT}}$ & 21.6 & 3.98 & 22.57 & 13.28 & 23.7 & 43.2 & 19.5 & 10.98 \\
EDU$_{\text{BGB}}$ & 14.3 & 7.48 & 27.96 & 17.92 & 14.7 & \textbf{83.5} & 68.8 & 16.76 \\
EDU$_{\text{GoG}}$ & 9.1 & \textbf{8.47} & \textbf{36.22} & \textbf{22.35} & 9.3 & 52.5 & 44.2 & 15.93 \\
EDU$_{\text{BoB}}$ & \textbf{7.8} & 8.40 & 36.08 & 22.24 & \textbf{7.3} & 83.4 & \textbf{76.1} & \textbf{17.18} \\ \bottomrule
\end{tabular}
\end{table}

\begin{table}[t]
\centering
\captionsetup{width=.665\textwidth}
\caption{Human evaluation results of PersonaChat, including response quality (left) and persona consistency (right). $\mathcal K$ is the Fleiss' kappa. The best results are in bold.}
\label{tab:dense_human}
\begin{tabular}{@{}l|rrr|rlrrl@{}}
\toprule
\textbf{Model} & \textbf{Flue.} & \textbf{Info.} & \textbf{Relv.} & \textbf{Per.C.} & \multicolumn{1}{c}{\textbf{+1}} & \multicolumn{1}{c}{\textbf{0}} & \multicolumn{1}{c}{\textbf{-1}} & \multicolumn{1}{c}{$\mathcal{K}$} \\ \midrule
Transformer~\cite{vaswani2017attention} & 3.05 & 2.57 & 2.72 & 0.05 & 30\% & 45\% & 25\% & 0.618 \\
CMAML~\cite{song-etal-2020-learning} & 3.36 & 2.40 & 3.09 & 0.24 & 34\% & 56\% & 10\% & 0.695 \\
GDR~\cite{song-etal-2020-generate} & 3.38 & 2.74 & 3.13 & 0.21 & 33\% & 55\% & 12\% & 0.669 \\
LIC~\cite{golovanov2020lost} & 3.70 & 3.53 & 3.47 & 0.39 & 55\% & 29\% & 16\% & 0.701 \\
GPT-2~\cite{radford2019language} & 3.79 & 3.22 & 3.79 & 0.47 & 61\% & 25\% & 14\% & 0.683 \\ \midrule
EDU$_{\text{ToT}}$ & 3.28 & 2.83 & 3.17 & 0.22 & 33\% & 56\% & 11\% & 0.615 \\
EDU$_{\text{BGB}}$ & 4.19 & 3.42 & 3.59 & \textbf{0.61} & 68\% & 25\% & 7\% & 0.477 \\
EDU$_{\text{GoG}}$ & \textbf{4.23} & \textbf{4.10} & 3.96 & 0.54 & 67\% & 20\% & 13\% & 0.518 \\
EDU$_{\text{BoB}}$ & 4.12 & 4.03 & \textbf{4.09} & 0.60 & 69\% & 22\% & 9\% & 0.734 \\ \bottomrule
\end{tabular}
\end{table}

\subsubsection{Results of Full PersonaChat}
\label{sec:dense-full}
We first report the automatic evaluation results on the PersonaChat dataset in Table~\ref{tab:dense_automatic}. As discussed in the evaluation metrics section (\cref{sec:exp_metrics}), we focus on two aspects, including response quality and persona consistency. We conduct the comparison between the proposed framework with different initialization schemes (detailed in \cref{sec:model_init}) and baseline models. 
From Table~\ref{tab:dense_automatic}, we can find that: 
1) Both the best response quality and persona consistency results are obtained by the proposed EDU framework, among which the EDU$_{\text{GoG}}$ achieves better response diversity, and the EDU$_{\text{BoB}}$ performs better at consistency understanding. All the improvements over the baseline models are statistically significant (two-tailed t-test, p-value < 0.01).
2) Comparing the baseline Transformer with our EDU$_{\text{ToT}}$ (Transformer-over-Transformer) model, both of which are trained from scratch, the EDU$_{\text{ToT}}$ has obtained significant improvements over the Transformer baseline on all metrics (two-tailed t-test, p-value < 0.01). A similar phenomenon can also be observed between the GPT-2 and the EDU$_{\text{GoG}}$. Although the EDU$_{\text{GoG}}$ is initialized from a smaller GPT, it still achieves better performance within the EDU framework. These results again demonstrate the effectiveness of the proposed framework on the personalized dialogue generation task.
3) By comparing different schemes within the EDU framework, we can find that initializing from the pre-trained language models consistently yields better results, but different types of language models will bring different benefits. The masked language model BERT can bring better results on consistency understanding, while the auto-regressive GPT is more suitable for generation. 
It concurs with our intuition that bidirectional attention can capture more information than the unidirectional attention for understanding.

Table~\ref{tab:dense_human} shows the human evaluation results on the full PersonaChat dataset. Generally, the human evaluation results are consistent with those of automatic metrics. Thus we take a closer look at the persona consistency through human annotations. We report the detailed portions of entailment (+1), irrelevant (0), and contradiction (-1), along with the Fleiss' kappa to measure inter-rater agreements in Table~\ref{tab:dense_human}. A high consistency score requires not only a high proportion of entailment but also a low proportion of contradiction.
The EDU framework reduces the proportion of contradictions while leveraging persona information as much as possible, resulting in the best consistency score.

\begin{table}[t]
\centering
\captionsetup{width=.77\textwidth}
\caption{Automatic evaluation results of EDU$_{\text{BoB}}$ under the low-resource PersonaChat setting, including response quality (left) and persona consistency (right). The ``$\dagger$'' means the minimum amount of data our model needed to outperform baselines' best results.}
\label{tab:dense-low-resource-auto}
\begin{tabular}{@{}l|llll|llll@{}}
\toprule
{\textbf{Model Setting}} & \textbf{PPL} & \textbf{Dist.1} & \textbf{Dist.2} & \textbf{D.AVG} & \textbf{p.Ent} & \textbf{p.Ctd} & \textbf{$\Delta$P} & \textbf{C.Score} \\ \midrule
Baselines' Best & 14.4 & 7.29 & 28.99 & 17.71 & 12.0 & 37.5 & 12.6 & 15.88 \\ \midrule
EDU$_{\text{BoB}}$ 1/8 Data & 11.6$^\dagger$ & 7.49$^\dagger$ & 27.10 & 17.30 & 11.3$^\dagger$ & 83.6$^\dagger$ & 72.3$^\dagger$ & 15.87 \\
EDU$_{\text{BoB}}$ 1/4 Data & 9.7 & 7.97 & 30.20$^\dagger$ & 19.09$^\dagger$ & 11.8 & 85.8 & 74.0 & 16.04$^\dagger$ \\
EDU$_{\text{BoB}}$ 1/2 Data & 8.9 & 8.13 & 33.08 & 20.61 & 8.1 & 81.9 & 73.8 & 16.36 \\ \bottomrule
\end{tabular}
\end{table}

\begin{table}[t]
\centering
\captionsetup{width=.67\textwidth}
\caption{Human evaluation results of EDU$_{\text{BoB}}$ under the low-resource PersonaChat setting, including response quality (left) and persona consistency (right). The ``$\dagger$'' means the minimum amount of data our model needs to outperform baselines' best results.}
\label{tab:dense-low-resourcen-human}
\begin{tabular}{@{}l|lll|lllll@{}}
\toprule
{\textbf{Model Setting}} & \textbf{Flue.} & \textbf{Info.} & \textbf{Relv.} & \textbf{Per.C.} & \textbf{+1} & \textbf{0} & \textbf{-1} & $\mathcal{K}$ \\ \midrule
Baselines' Best & 3.79 & 3.53 & 3.79 & 0.47 & 61\% & 25\% & 14\% & 0.683 \\ \midrule
EDU$_{\text{BoB}}$ 1/8 Data & 4.17$^\dagger$ & 3.48 & 4.12$^\dagger$ & 0.62$^\dagger$ & 66\% & 30\% & 4\% & 0.551 \\
EDU$_{\text{BoB}}$ 1/4 Data & 4.19 & 3.47 & 4.17 & 0.60 & 66\% & 28\% & 6\% & 0.497 \\
EDU$_{\text{BoB}}$ 1/2 Data & 4.03 & 3.70$^\dagger$ & 3.94 & 0.61 & 68\% & 25\% & 7\% & 0.529 \\ \bottomrule
\end{tabular}
\end{table}

\begin{table}[ht]
\centering
\captionsetup{width=0.81\textwidth}
\caption{Automatic evaluation results of other initialization schemes from full PersonaChat to low-resource settings. The metrics evaluate response quality (left) and persona consistency (right). }
\label{tab:dense-low-resource framework}
\begin{tabular}{@{}c|c|rrrr|rrrr@{}}
\toprule
\textbf{Model} & \multicolumn{1}{l|}{\textbf{Data Setting}} & \textbf{PPL} & \textbf{Dist.1} & \textbf{Dist.2} & \textbf{D.AVG} & \textbf{p.Ent} & \textbf{p.Ctd} & \textbf{$\Delta$P} & \textbf{C.Score} \\ \midrule
\multirow{4}{*}{EDU$_{\text{ToT}}$} & Full Data & 21.6 & 3.98 & 22.57 & 13.28 & 23.7 & 43.2 & 19.5 & 10.98 \\
 & 1/2 Data & 29.9 & 3.11 & 18.89 & 11.00 & 33.6 & 52.5 & 18.9 & 9.63 \\
 & 1/4 Data & 38.6 & 2.52 & 15.33 & 8.93 & 40.2 & 55.7 & 15.5 & 9.08  \\
 & 1/8 Data & 45.2 & 2.03 & 10.93 & 6.48 & 46.8 & 57.5 & 10.7 & 8.04  \\ \midrule
\multirow{4}{*}{EDU$_{\text{BGB}}$} & Full Data & 14.3 & 7.48 & 27.96 & 17.92 & 14.7 & 83.5 & 68.8 & 16.76 \\
 & 1/2 Data & 14.9 & 7.24 & 27.37 & 17.31 & 15.9 & 82.3 & 66.4 & 16.08 \\
 & 1/4 Data & 15.4 & 7.13 & 27.41 & 17.27 & 17.0 & 84.1 & 67.1 & 15.89 \\
 & 1/8 Data & 16.7 & 6.89 & 26.81 & 16.85 & 16.9 & 83.4 & 66.5 & 15.76 \\ \midrule
\multirow{4}{*}{EDU$_{\text{GoG}}$} & Full Data & 9.1 & 8.47 & 36.22 & 22.35 & 9.3 & 52.5 & 44.2 & 15.93 \\
 & 1/2 Data & 9.7 & 7.88 & 36.03 & 21.96 & 9.9 & 55.2 & 45.3 & 15.68 \\
 & 1/4 Data & 10.0 & 7.83 & 36.12 & 21.97 & 10.3 & 54.1 & 43.8 & 15.39 \\
 & 1/8 Data & 10.7 & 7.49 & 34.93 & 21.21 & 10.9 & 55.4 & 44.5 & 15.38 \\ \bottomrule
\end{tabular}
\end{table}

\subsubsection{Results of Less Personalized Data}
\label{sec:dense-low-resource}
Since we have observed improvements with a large margin on the full PersonaChat dataset, we further evaluate the EDU framework under a low-resource scenario, where we gradually reduce the number of training examples by halving.
According to the performance on the full PersonaChat evaluations, we first report the automatic and human evaluation results of EDU$_{\text{BoB}}$'s low-resource performance in Table~\ref{tab:dense-low-resource-auto} and Table~\ref{tab:dense-low-resourcen-human}, respectively.
From the evaluation results, we can see that:
1) The EDU$_{\text{BoB}}$ can outperform most of the baselines' best results even by using only 1/8 of the training data. 
2) Due to the EDU framework regularizing generation with understanding and learning consistency understanding from non-dialogue inference dataset, the EDU$_{\text{BoB}}$ model presents a stable yet good performance on consistency-related metrics regardless of the size of the dialogue set.
3) The low perplexity indicates that the EDU$_{\text{BoB}}$ model inherits a strong language model from the pre-trained BERT. And thus, the EDU$_{\text{BoB}}$ model can focus on the learning of alignment between personas and responses from the limited training data.

We further report the low-resource automatic evaluation results of three other framework initialization schemes in Table~\ref{tab:dense-low-resource framework} to have a thorough assessment of the stack-propagation framework. From Table~\ref{tab:dense-low-resource framework} we can see that:
1) Both response generation and consistency understanding can benefit from a good initialization. Different from the EDU$_{\text{BGB}}$ and EDU$_{\text{GoG}}$, both the generation and understanding performance of EDU$_{\text{ToT}}$ decreases significantly with the decrease of data amount. A plausible reason is that the semantic representation is the basis of both tasks, where under the low-resource setting, the from-scratch trained EDU$_{\text{ToT}}$ model can not learn well.
2) Framework initialized with auto-regressive GPT model, i.e., the EDU$_{\text{GoG}}$, constantly performs better at response quality metrics, while the BERT model can be a better regularizer for consistency understanding tasks.
3) Initializing the framework with the same type of language models have a positive effect on the perplexity. Both EDU$_{\text{BoB}}$ and EDU$_{\text{GoG}}$ can achieve a perplexity lower than 10, while the perplexities of the EDU$_{\text{BGB}}$ model are significantly higher. It is reasonable since BERT and GPT do not have an originally shared embedding layer, which will be learned from scratch during training.

\begin{table}[t]
\centering
\captionsetup{width=.51\textwidth}
\caption{Perplexity comparison on PersonaChat among pre-trained dialogue models, along with the number of parameters.}
\label{tab:dense-analysis-ppl}
\begin{tabular}{@{}p{0.3\textwidth}rr@{}}
\toprule
\textbf{Pre-trained Dialogue Model} & \textbf{PPL} & \textbf{\# Parameters} \\ \midrule
GPT~\cite{radford2018improving} & 17.6 & 110 M  \\
BERT~\cite{devlin-etal-2019-bert} & 25.7 & 110 M  \\
% TransferTransfo~\cite{wolf2019transfertransfo} & 17.5 & 124 M  \\
GPT-2~\cite{radford2019language} & 14.4 & 345 M \\
Adapter Bot~\cite{lin2021adapter} & 11.1 & 350 M \\
BART~\cite{lewis2020bart} & 10.0 & 406 M \\
% DDD~\cite{shuster2020dialogue} & 10.8 & -  \\
BST Generative~\cite{roller2021recipes} & 8.4 & 2700 M \\ \midrule
EDU$_{\text{GoG}}$  & 9.1 & 340 M \\
EDU$_{\text{BoB}}$  & 7.8 & 340 M \\ \bottomrule
\end{tabular}
\end{table}

\begin{table}[t]
\centering
\captionsetup{width=.85\textwidth}
\caption{Comparisons of automatic consistency score (C.Score) and averaged distinct-1 and distinct-2 (D.AVG) between the generative (Gen.) and re-ranking (ReRank) way. The C.Score is from an NLI model trained on the DNLI dataset, and the \textit{oracle} denotes re-ranking with the model trained on DNLI dataset.}
\label{tab:rerank}
\begin{tabular}{@{}l|ccc|ccc@{}}
\toprule
\multirow{2}{*}{\textbf{Model}} & \multicolumn{3}{c|}{\textbf{C.Score}} & \multicolumn{3}{c}{\textbf{D.AVG}} \\ \cmidrule(l){2-7} 
 & \multicolumn{1}{c}{\textbf{Gen.}} & \multicolumn{1}{c}{\textbf{ReRank$_\text{mnli}$}} & \multicolumn{1}{c|}{\textbf{ReRank$_\text{oracle}$}} & \multicolumn{1}{c}{\textbf{Gen.}} & \multicolumn{1}{c}{\textbf{ReRank$_\text{mnli}$}} & \multicolumn{1}{c}{\textbf{ReRank$_\text{oracle}$}} \\ \midrule
Transformer~\cite{vaswani2017attention} & 1.20 & 3.17 & 9.34  & 10.47 & 10.42 & 10.03 \\
LIC~\cite{golovanov2020lost} & 14.12 & 14.53 & 17.97 & 17.64 & 16.73 & 17.89 \\
GPT-2~\cite{radford2019language} & 15.88 & 16.38 & 18.23 & 17.71 & 18.38 & 17.31 \\ \midrule
E+D & 15.97 & 16.45 & 17.12 & 17.37 & 17.35 & 16.82  \\
EDU$_{\text{BoB}}$ & 17.18 & 17.20 & 18.81 & 22.24 & 21.82  & 22.51 \\ \bottomrule
\end{tabular}
\end{table}

\subsubsection{Analysis of Framework Performance}
\label{sec:dense-analysis}

\paragraph{Comparison on Perplexity} Since the proposed framework achieves a remarkably low perplexity, especially for the two homologous initialized models, i.e., EDU$_{\text{BoB}}$ and EDU$_{\text{GoG}}$, we further conduct a comparison on the perplexity with more pre-trained dialogue models. We report the perplexity comparison results and the number of model parameters in Table~\ref{tab:dense-analysis-ppl}.
As we can see, the EDU framework achieves competitive perplexity compared with much larger model. Moreover, it also achieves much better perplexity performance than the initializing language models, demonstrating the effectiveness of the proposed stack-propagation framework. 
% Detailed ablation analysis will be provided in the following section.

\paragraph{Comparison with Re-Ranking}
The stack-propagation framework leverages language inference data with the unlikelihood objective.
However, baseline models do not have such capability. On the other hand, compared with complex architecture design, a more straightforward way to leverage language inference data is re-ranking. Specifically, an NLI model will be trained on the inference data, and the candidate responses will be re-ranked according to the NLI model's entailment score. We apply a beam search strategy with a size of 32 to the evaluated models to get their top-20 responses. Here we trained two NLI models for re-ranking: one is on the MNLI dataset, which is the inference data used in the EDU framework, and another is on the DNLI dataset, which is also used for evaluation and thus noted as {\it oracle}. We report the comparison results in Table~\ref{tab:rerank}.
As we can see from the C.Score, simply re-ranking with MNLI trained model only improves a little consistency score. In contrast, the oracle model improves consistency significantly, which is within our expectations. The results prove that the re-ranking is a feasible scheme for improving consistency as long as there are enough annotated data. We also remove $\mathbb U$ and directly re-ranking the responses from $\mathbb D$. Results show that the generative manner gets a consistency score of 17.18. In contrast, the re-ranking manner only gets 16.45, proving the effectiveness of leveraging consistency understanding as a regularizer for generation in a joint framework.
On the other hand, the re-ranking strategy does not contribute to response quality. From the D.AVG results we can see that the distinct metrics of all the re-ranked results are not much different whether using MNLI or DNLI for re-ranking. Under the re-ranking scenario, the dialogue generation models are independent of the re-ranking model. This disconnection means that the data for re-ranking training does not contribute to the generation model's optimization, which leads to the loss of learning diverse expressions from these additional data.

\begin{table}[t]
\centering
\captionsetup{width=.89\textwidth}
\caption{EDU$_{\text{BoB}}$ ablation results of both automatic metrics and human evaluations on full PersonaChat dataset.}
\label{tab:dense-ablation-bob}
\begin{tabular}{@{}l|rrrr|rrrrrrr@{}}
\toprule
\multirow{2}{*}{\textbf{Ablation}} & \multicolumn{4}{c|}{\textbf{Objective}} & \multicolumn{7}{c}{\textbf{Subjective}} \\ \cmidrule(l){2-12} 
 & \textbf{PPL} & \textbf{D.AVG} & \textbf{C.Score} & \textbf{$\Delta$P} & \multicolumn{1}{l}{\textbf{Flue.}} & \multicolumn{1}{l}{\textbf{Info.}} & \multicolumn{1}{l}{\textbf{Relv.}} & \multicolumn{1}{l}{\textbf{Per.C.}} & \textbf{+1} & \textbf{0} & \textbf{-1} \\ \midrule
EDU$_{\text{BoB}}$ & 7.8 & 22.24 & 17.18 & 76.1 & 4.12 & 4.03 & 4.09 & 0.60 & 69\% & 22\% & 9\%  \\ \midrule
w/o UL & 8.1 & 17.89 & 16.21 & 7.8 & 3.81 & 3.50 & 3.80 & 0.48 & 64\% & 20\% & 16\%   \\
E+D & 23.6 & 17.37 & 15.97 & 4.9 & 3.65 & 3.18 & 3.60 & 0.45 & 62\% & 21\% & 17\%   \\
E & 25.7 & 17.41 & 15.76 & 7.1 & 3.69 & 3.28 & 3.60 & 0.42 & 58\% & 26\% & 16\%   \\ \bottomrule
\end{tabular}
\end{table}

\begin{table}[t]
\captionsetup{width=.89\textwidth}
\caption{EDU$_{\text{GoG}}$ ablation results of both automatic metrics and human evaluations on full PersonaChat dataset.}
\label{tab:dense-ablation-gog}
\begin{tabular}{@{}l|rrrr|rrrrrrr@{}}
\toprule
\multirow{2}{*}{\textbf{Ablation}} & \multicolumn{4}{c|}{\textbf{Objective}} & \multicolumn{7}{c}{\textbf{Subjective}} \\ \cmidrule(l){2-12} 
 & \textbf{PPL} & \textbf{D.AVG} & \textbf{C.Score} & \textbf{$\Delta$P} & \multicolumn{1}{l}{\textbf{Flue.}} & \multicolumn{1}{l}{\textbf{Info.}} & \multicolumn{1}{l}{\textbf{Relv.}} & \multicolumn{1}{l}{\textbf{Per.C.}} & \textbf{+1} & \textbf{0} & \textbf{-1} \\ \midrule
EDU$_{\text{GoG}}$ & 9.1 & 22.35 & 15.93 & 44.2 & 4.23 & 4.10 & 3.96 & 0.54 & 67\% & 20\% & 13\% \\ \midrule
w/o UL & 9.7 & 18.53 & 15.32 & 5.6 & 3.92 & 3.61 & 3.90 & 0.45 & 65\% & 15\% & 20\% \\
E+D & 22.8 & 17.18 & 15.29 & 4.9 & 3.25 & 3.05 & 3.46 & 0.44 & 66\% & 12\% & 22\% \\
E & 17.6 & 17.53 & 15.31 & 8.1 & 3.73 & 3.38 & 3.69 & 0.43 & 63\% & 17\% & 20\% \\ \bottomrule
\end{tabular}
\end{table}

\paragraph{Ablation Study}
After going through the EDU framework's performance in detail, there are still several questions remaining: 
1) from where does the regularizer $\mathbb{U}$ obtain the consistency understanding capability?
2) can the EDU models understand persona consistency just through training on the personalized dialogues?
3) does the good performance, such as the low PPL, come from the initialization of the pre-trained language model or the EDU framework?

In order to answer these questions, we ablate the two best performing schemes of the stack-propagation framework, i.e., EDU$_{\text{BoB}}$ and EDU$_{\text{GoG}}$, in the following ways: 
1) {\bf w/o UL}. This ablation only removes the unlikelihood objective in the regularizer $\mathbb{U}$. 
2) {\bf ${\mathbf E}$+${\mathbf{D}}$}. This ablation removes the unlikelihood objective and the regularizer $\mathbb U$.
3) {\bf ${\mathbf E}$}. This ablation removes the unlikelihood objective and both decoders. Thus it is the initialization model, i.e., BERT or GPT.

We report EDU$_{\text{BoB}}$'s ablation results on full PersonaChat in Table~\ref{tab:dense-ablation-bob} and EDU$_{\text{GoG}}$'s ablation results in Table~\ref{tab:dense-ablation-gog}.
From these ablation results, we can find:
1) The unlikelihood objective is the key to obtaining consistency understanding capability. In training, the unlikelihood objective assigns large perplexity to the contradictions. And thus, in the generation, the contradicted responses are less likely to be generated as they have larger losses.
And as shown in both tables, after removing the unlikelihood objective, all ablated models suffer from significant performance degradations in consistency metrics (Per.C. and $\Delta$P).
2) Models barely understand consistency if just training on personalized dialogues. 
Without the negative training (w/o UL), all the ablated models in both tables perform poorly on $\Delta$P, indicating they cannot distinguish contradiction from entailment. Although some of their Per.C. metrics still look good, it may come from just mimicking and copying words rather than understanding.
3) The good performance comes from the stack-propagation framework. 
There are significant performance gaps between the full EDU models and the initialization BERT and GPT. And among these metrics, the regularizer $\mathbb U$ contributes most to the perplexity, as the perplexity drops significantly after ablating $\mathbb U$. These results verify the necessity of the EDU framework's architecture.

\begin{figure}[t]
\centering
\includegraphics[width=.91\columnwidth]{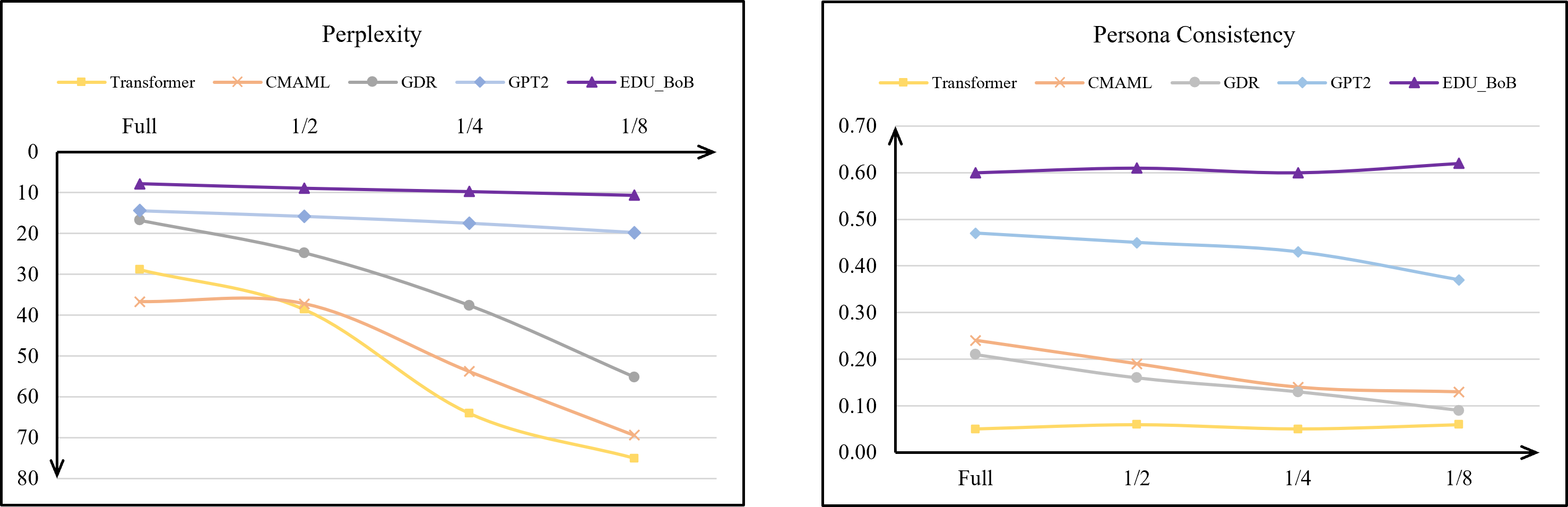}
\captionsetup{width=.91\textwidth}
\caption{ Illustrations of different models’ performance (better readability in color) under different amounts of personalized data (PersonaChat). The abscissa represents the amount of training data used. Left: the automatic metric perplexity, the lower, the better. Right: the human evaluation score of persona consistency, i.e., Per.C., the higher, the better.}
\label{fig:low_resource}
\end{figure}

\paragraph{Low-resource Performance Visualization}
To intuitively understand the stable performance of the EDU framework under the low-resource scenario, we illustrate the performance curves of the framework and baseline models under different amounts of personalized data in Figure~\ref{fig:low_resource}.
As we can see, under different data volumes, our method presents stable performances on both automatic perplexity and human-evaluated consistency scores.
For perplexity, with the decrease in the amount of data, the performances of non-pretrained methods show an apparent downward trend. It is reasonable because these non-pretrained methods could only learn the language modeling from the training data. And for the pre-trained models, they have already learned a robust language model from massive unlabeled text data, thus achieving a stable performance.
Similar phenomena can also be seen in the persona consistency. However, the pre-trained GPT-2 also shows a declining trend. An important reason is that the general pre-training procedure can not effectively endow models with the capability to understand the persona consistency. As a result, GPT-2 still needs to learn consistency from the personalized dialogue data.

\begin{figure}[t]
\centering
\includegraphics[width=.91\columnwidth]{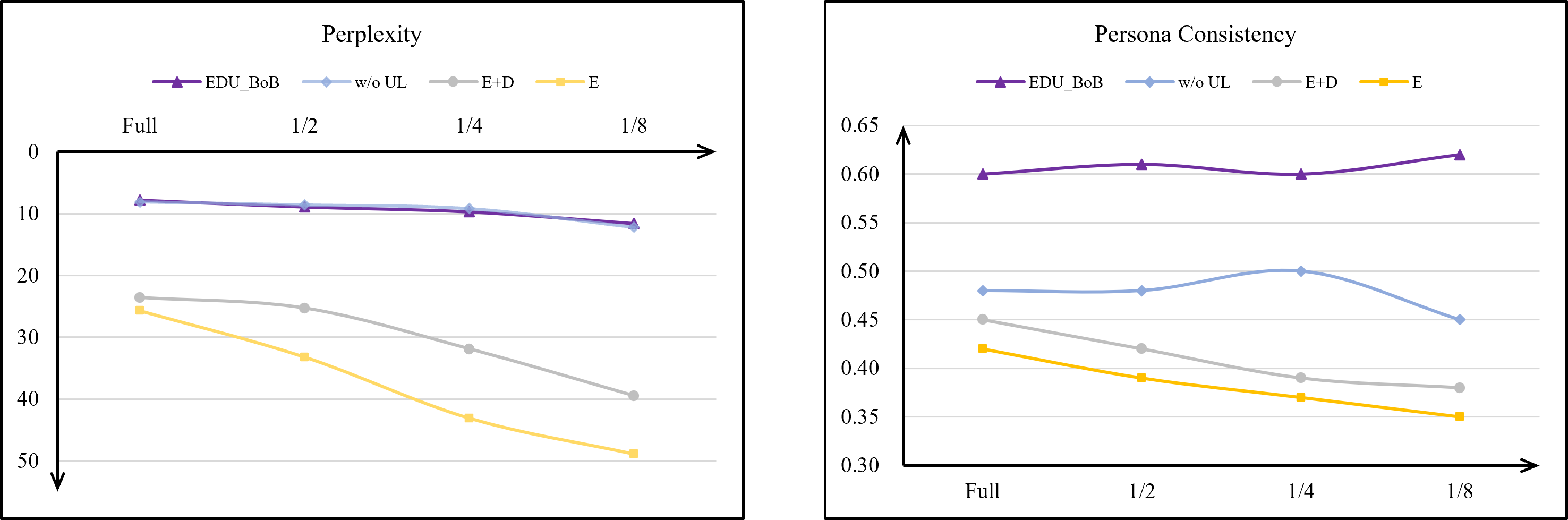}
\captionsetup{width=.91\textwidth}
\caption{Performance illustrations of EDU$_{\text{BoB}}$ and its ablations (better readability in color) under different personalized data amounts (PersonaChat). The abscissa represents the amount of training data used. Left: the automatic metric perplexity, the lower, the better. Right: the human-evaluated score of persona consistency, i.e., Per.C., the higher, the better.}
\label{fig:ablation}
\end{figure}

After verifying the stability of performance under low resource scenarios, an important question is where the stability comes from. 
To answer this question, we illustrate the performance of EDU$_{\text{BoB}}$ and its ablations under different amounts of personalized data in Figure~\ref{fig:ablation}. We can see that the performance of both {\bf ${\mathbf E}$} and  {\bf ${\mathbf E}$+${\mathbf{D}}$} decrease when there is fewer data available. 
In contrast, regularizer $\mathbb U$ plays an essential role in stabilizing performance under different data volumes. These results show that the stability of low-resource performance benefits from the stack-propagation architecture.

\begin{table}[t]
\centering
\captionsetup{width=.92\textwidth}
\caption{Results of training EDU framework with dialogue inference datasets, including the DNLI and DECODE.
% Both automatic metrics (Objective) and human evaluations (Subjective) are reported. 
The ``CON'' denotes the ratio of persona-contradicted responses. The CON$_\text{DNLI}$ and CON$_\text{DECODE}$ denote that the ratio comes from a detector model trained on the DNLI and DECODE. The CON$_\text{Human}$ denotes human-annotated contradiction ratio.
}
\label{tab:further:dnli-train}
\resizebox{.92\textwidth}{!}{%
\begin{tabular}{@{}l|c|cccc|cccc@{}}
\toprule
\multirow{2}{*}{\textbf{Model}} &
  \multirow{2}{*}{\textbf{\begin{tabular}[c]{@{}c@{}}Consistency\\ Data\end{tabular}}} &
  \multicolumn{4}{c|}{\textbf{Objective}} &
  \multicolumn{4}{c}{\textbf{Subjective}} \\ \cmidrule(l){3-10} 
 &
   &
  \textbf{PPL} &
  \textbf{D.AVG} &
  \textbf{CON$_\text{DNLI}$} &
  \textbf{CON$_\text{DECODE}$} &
  \textbf{Flue.} &
  \textbf{Info.} &
  \textbf{Relv.} &
  \textbf{CON$_\text{Human}$} \\ 
\midrule
\multirow{3}{*}{EDU$_{\text{GoG}}$} & MNLI   & 9.1  & \textbf{22.35} & 16.68\% & 15.36\% & \textbf{4.13} & \textbf{4.09} & 3.99 & 14\% \\
                                    & DNLI   & \textbf{8.9}  & 21.07 & 9.40\% & 11.79\% & 4.12 & 4.02 & \textbf{4.01} &  10\%    \\
                                    & DECODE & 9.0  & 21.98 & \textbf{9.24\%} & \textbf{6.63\%} & \textbf{4.13} & 4.03     & 3.98 &  \textbf{8\%}   \\
\midrule
\multirow{3}{*}{EDU$_{\text{BoB}}$} & MNLI   & \textbf{7.8} & \textbf{22.24} & 12.99\% & 14.52\% & 4.05 & \textbf{4.01} & \textbf{4.10} & 11\%  \\
                                    & DNLI   & 8.1 & 20.31 & \textbf{7.52\%} & 11.79\% & \textbf{4.07} &  3.92 & 4.04 & 9\% \\
                                    & DECODE & 7.9 & 21.65 & 8.81\% & \textbf{6.41\%} & 4.06 &  3.94  & 4.07 & \textbf{7\%} \\
\bottomrule
\end{tabular}
}%
\end{table}

\begin{table}[t]
\centering
\captionsetup{width=.92\textwidth}
\caption{Comparison results with improved baselines, including BART$_\text{large}$ and T5$_\text{large}$. Accordingly, the EDU framework is also initialized from stronger models.
The human-annotated cases are re-sampled and thus different from the main results.
}
\label{tab:further:improved_baseline}
\resizebox{.92\textwidth}{!}{%
\begin{tabular}{@{}l|c|ccccc|cccc@{}}
\toprule
\multirow{2}{*}{\textbf{Architecture}} & \multirow{2}{*}{\textbf{Model}} & \multicolumn{5}{c|}{\textbf{Objective}} & \multicolumn{4}{c}{\textbf{Subjective}} \\ \cmidrule(l){3-11} 
 &  & \textbf{PPL} & \textbf{Dist.1} & \textbf{Dist.2} & \textbf{C.Score} & \textbf{$\Delta$P} & \textbf{Flue.} & \textbf{Info.} & \textbf{Relv.} & \textbf{Per.C.} \\ \midrule
BART~\cite{lewis2020bart} & bart-large & 10.0 & 8.32 & 38.91 & 16.97 & 12.4 & 4.19 & 4.22 & 4.15 & 0.54 \\
T5~\cite{T5} & t5-large & 7.5 & 8.69 & 39.50 & 17.51 & 17.3 & 4.27 & 4.26 & 4.21 & 0.58 \\
\midrule
\multirow{2}{*}{EDU$_{\text{GoG}}$} & openai-gpt & 9.1 & 8.47 & 36.22 & 15.93 & 44.2 & 4.13 & 4.09 & 3.99 & 0.52 \\
 & gpt2-medium & 7.6 & 8.77 & \textbf{43.08} & 17.24 & 50.6 & \textbf{4.32} & 4.25 & 4.17 & 0.54 \\
\midrule
\multirow{3}{*}{EDU$_{\text{BoB}}$} & bert-base & 7.8 & 8.40 & 36.08 & 17.18 & 76.1 & 4.05 & 4.01 & 4.10 & 0.59 \\
 & roberta-base & 6.6 & 8.64 & 41.79 & 17.86 & 79.9 & 4.27 & 4.18 & \textbf{4.23} & 0.60  \\
 & bert-large & \textbf{4.9} & \textbf{8.93} & 42.75 & \textbf{18.10} & \textbf{81.7} & 4.29 & \textbf{4.31} & 4.22 & \textbf{0.63} \\ 
\bottomrule
\end{tabular}
}%
\end{table}

\subsubsection{Further Discussion}
\label{sec:dense:further-explore}
\paragraph{Training with Dialogue Inference Data.}
The EDU framework is designed to gain the dialogue understanding capability from non-dialogue inference data, such as the MNLI.
After verifying the effectiveness of training with the non-dialogue inference data, here we aim to investigate whether the high-quality dialogue inference datasets, such as the DNLI~\cite{WelleckDNLI} and DECODE~\cite{nie2021dolphin}, can help the EDU framework learn better understanding capability.
Specifically, when calculating the unlikelihood loss in equation~\ref{formula:unlikelihood_sum}, here we ignore the low-resource constraints and use positive and negative examples from the dialogue inference datasets DNLI and DECODE. 
For DNLI, the positive and negative examples are from its entailment and contradiction categories. For DECODE, since it only has contradiction and non-contradiction categories, the negative examples are from its aggregated contradiction instances, and the positive examples are sampled from other dialogues. For the statistics of the two datasets, please refer to Table~\ref{tab:inference_datasets}. 
We train the \textit{dialogue contradiction detectors}~\citep{nie2021dolphin} from a RoBERTa model on both DNLI and DECODE datasets. The two detectors indicate the contradiction ratios in the generated responses and cross-validate each other's effectiveness.
For human evaluation, the same as the main experiments, we sample 100 test cases and collect the responses from each setting for human annotation.

We report the results in Table~\ref{tab:further:dnli-train}. The main findings are as follows:
1) From both model predictions and human annotations, the proportion of contradictions in the DNLI and DECODE trained models is lower than in the MNLI trained models, indicating that the low-resource EDU framework can also benefit from high-quality data.
2) The high-quality dialogue understanding data does not improve the quality of dialogue responses. It even negatively influences the diversity of the response, which can be attributed to the limited scales and patterns compared with the non-dialogue MNLI dataset.
3) We can observe that the model-based contradiction detection has a bias towards the models trained with the same data. This phenomenon highlights the importance of cross-validation, i.e., a dialogue model trained on the DNLI dataset should be evaluated with models trained on other datasets, such as the DECODE.

\paragraph{Comparisons with Improved Baselines.}
Considering previous baseline models are primarily built upon the GPT and GPT-2 models, here we further compare the EDU framework with two strong encoder-decoder pre-trained models, including the BART~\cite{lewis2020bart} and T5~\citep{T5}.
We set up new baselines with their \textit{large} size models.
To mitigate the performance gaps in pre-training, we also initialize the EDU framework with stronger language models, including the GPT-2$_\text{medium}$, RoBERTa$_\text{base}$, and BERT$_\text{large}$.
For the configurations of the three initializing models, please refer to Table~\ref{tab:model_config}.

We report the experimental results in Table~\ref{tab:further:improved_baseline}. As we can see, the BART and T5 models achieve impressive performances on both automatic metrics and human evaluations. Nevertheless, when initialized with comparable language models, the EDU framework catches up with the strong baselines on all metrics. Although the RoBERTa$_\text{base}$ is of the same size as the BERT$_\text{base}$, the RoBERTa-initialized model still outperforms the BERT model due to the better approach of pre-training~\cite{liu2019roberta}.
On the other hand, the results in Table~\ref{tab:further:improved_baseline} also show that stronger language models mainly improve the quality of dialogue responses, but the improvement of consistency is relatively limited.

\begin{table}[t]
\centering
\captionsetup{width=.87\textwidth}
\caption{Automatic metrics (left) and human evaluation results (right) on the random testset of PersonalDialog. 
Most dialogues in the random testset are persona-irrelevant. The best results are in bold.
}
\label{tab:sparse-random}
\begin{tabular}{@{}p{.2\textwidth}|rrr|rrrrrrr@{}}
\toprule
\textbf{Model} & \textbf{PPL} & \textbf{D.AVG} & \textbf{C.Score} & \textbf{Flue.} & \textbf{Info.} & \textbf{Relv.} & \textbf{Per.C.} & \textbf{+1} & \textbf{0} & \textbf{-1} \\ \midrule
Transformer~\cite{vaswani2017attention} & 43.7 & 13.27 & 0.95 & 3.26 & 2.38 & 2.72 & 0.00 & 4\% & 92\% & 4\% \\
LIC~\cite{golovanov2020lost} & 47.8 & 17.33 & 4.08 & 3.68 & 2.66 & 2.92 & 0.02 & 4\% & 94\% & 2\% \\
AttentionRouting~\cite{Zheng_Zhang_Huang_Mao_2020} & 34.2 & 18.46 & -2.14 & 3.71 & 2.58 & \textbf{3.02} & -0.03 & 0\% & 97\% & 3\%  \\ \midrule
EDU$_{\text{ToT}}$ & 39.4 & 15.75 & 0.39 & 3.35 & 2.43 & 2.86 & -0.01 & 1\% & 97\% & 2\%  \\
EDU$_{\text{BGB}}$ & 29.2 & 18.65 & \textbf{4.26} & 3.41 & 2.57 & 2.91 & \textbf{0.03} & 3\% & 97\% & 0\% \\
EDU$_{\text{GoG}}$ & 22.1 & \textbf{19.52} & 0.03 & 3.74 & \textbf{2.74} & 2.99 & -0.01 & 2\% & 95\% & 3\% \\
EDU$_{\text{BoB}}$ & \textbf{18.5} & 18.87 & 2.10 & \textbf{3.75} & 2.69 & 2.98 & 0.01 & 3\% & 95\% & 2\% \\ \bottomrule
\end{tabular}
\end{table}

\subsection{Validation Results on the Persona-Sparse Scenario}
\label{sec:exp_sparse_results}

We further validate the stack-propagation framework on a persona-sparse scenario with the PersonalDialog dataset.
Persona-sparse is a typical scenario for the data collected from social media: although there are some publicly available user profiles, most users would not exhibit their persona exhaustively. Therefore, the personalized dialogue data collected in this way are usually very sparse in persona information.

To have a more intuitive understanding of the ``sparsity'' in the PersonalDialog dataset, the same annotation team for the PersonalDialog experiments is also recruited to annotate whether the ground-truth response is persona-relevant. Results show that only 1\% of responses are persona-relevant in the random testset and 28\% in the biased testset. We calculate the Fleiss’ kappa among the five annotators and obtain a kappa of 0.774, which means {\it substantial agreement}.
We first report the evaluation results on the random testset in Table~\ref{tab:sparse-random}.
On the random test set, experimental results demonstrate that the EDU framework still gains advantages over baseline models, but the improvements on consistency-related metrics are not as significant as on PersonaChat. It seems reasonable as the task has turned into the ordinary dialogue generation on the random testset, and all the compared methods are at the same language model scale. As a result, our framework's advantages on exploiting consistency understanding can not be effectively exhibited.
Evidence can be found in the human-annotated consistency results: about 95\% of the generated responses in all methods are persona-irrelevant, in sharp contrast to the results of the persona-dense PersonaChat dataset.

Table~\ref{tab:sparse-biased} reports the evaluation results on the biased testset. Unlike the random testset, a good portion of dialogues in the biased testset is persona-relevant. This setting verifies whether the model can effectively learn from large-scale persona-sparse datasets and effectively generate persona-based dialogues.
From the results, we can see that:
1) The proposed EDU framework achieves the best results on all metrics.
2) The good performances on C.Score and Per.C. indicate that the EDU framework can be effectively trained from a dataset with limited personalized dialogues.
3) The human-annotated consistency results (+1/0/-1) confirm the difference between the biased testset and the previous random testset, where the results on the biased testset tend to be more persona-relevant.

\begin{table}[t]
\centering
\captionsetup{width=\textwidth}
\caption{Automatic metrics (left) and human evaluation results (right) on the biased testset of PersonalDialog. 
A considerable number of dialogues in the biased testset are persona-relevant. The best results are in bold.
}
\label{tab:sparse-biased}
\resizebox{\textwidth}{!}{%
\begin{tabular}{@{}p{.2\textwidth}|rrrr|rrrrrrr@{}}
\toprule
\textbf{Model} & \textbf{PPL} & \textbf{D.AVG} & \textbf{$\Delta$P} & \textbf{C.Score} & \textbf{Flue.} & \textbf{Info.} & \textbf{Relv.} & \textbf{Per.C.} & \textbf{+1} & \textbf{0} & \textbf{-1} \\ \midrule
Transformer~\cite{vaswani2017attention} & 83.2 & 20.32 & 3.28 & 1.04 & 3.54 & 2.58 & 2.84 & 0.03 & 19\% & 65\% & 16\% \\
LIC~\cite{golovanov2020lost} & 43.3 & 23.81 & 2.86 & 8.25 & 3.72 & 3.01 & 3.04 & 0.08 & 18\% & 72\% & 10\% \\
AttentionRouting~\cite{Zheng_Zhang_Huang_Mao_2020} & 38.7 & 25.67 & 3.08 & 11.72 & 3.78 & 3.11 & 3.10 & 0.13 & 26\% & 61\% & 13\% \\ \midrule
EDU$_{\text{ToT}}$ & 53.2 & 21.98 & 23.52 & 1.67 & 3.57 & 2.89 & 2.85 & 0.10 & 25\% & 60\% & 15\%  \\
EDU$_{\text{BGB}}$ & 29.9 & 25.34 & 79.65 & 12.52 & 3.75 & 2.98 & 2.99 & 0.14 & 24\% & 66\% & 10\% \\
EDU$_{\text{GoG}}$ & 23.6 & \textbf{27.86} & 45.78 & 11.94 & \textbf{3.88} & \textbf{3.25} & 3.16 & 0.13 & 27\% & 59\% & 14\% \\
EDU$_{\text{BoB}}$ & \textbf{19.5} & 26.84 & \textbf{85.40} & \textbf{12.76} & 3.84 & 3.13 & \textbf{3.17} & \textbf{0.15} & 24\% & 67\% & 9\% \\ \bottomrule
\end{tabular}
}
\end{table}

\begin{table}[t]
\centering
\captionsetup{width=\textwidth}
\caption{Ablation results of the EDU$_{\text{BoB}}$ model on both random and biased testsets of the PersonalDialog, along with a data ablation (w/o persona) to investigate whether the persona matters under two different testsets.}
\label{tab:sparse-bob-ablation}
\resizebox{\textwidth}{!}{%
\begin{tabular}{@{}l|rrrrrr|rrrrrr|r@{}}
\toprule
\multirow{2}{*}{\textbf{Ablation}} & \multicolumn{6}{c|}{\textbf{Random Testset}} & \multicolumn{6}{c|}{\textbf{Biased Testset}} & \multicolumn{1}{c}{\textbf{KvPI}} \\ \cmidrule(l){2-14} 
 & \textbf{PPL} & \textbf{C.Score} & \textbf{Flue.} & \textbf{Info.} & \textbf{Relv.} & \textbf{Per.C.} & \textbf{PPL} & \textbf{C.Score} & \textbf{Flue.} & \textbf{Info.} & \textbf{Relv.} & \textbf{Per.C.} & \textbf{$\Delta$P} \\ \midrule
EDU$_{\text{BoB}}$ & 18.5 & 2.10 & 3.75 & 2.69 & 2.98 & 0.01 & 19.5 & 12.76 & 3.84 & 3.13 & 3.17 & 0.15 & 85.40 \\ \midrule
w/o persona & 18.3 & 2.08 & 3.73 & 2.70 & 3.02 & 0.01 & 25.6 & 4.03 & 3.79 & 2.98 & 2.97 & 0.05 & 4.58 \\ \midrule
w/o UL & 19.3 & -3.13 & 3.73 & 2.57 & 2.93 & -0.06 & 20.1 & 10.53 & 3.79 & 2.92 & 3.10 & 0.09 & 4.10 \\
E+D & 31.7 & 0.15 & 3.74 & 2.68 & 2.96 & -0.01 & 38.0 & 9.75 & 3.74 & 3.15 & 3.06 & 0.08 & 2.80 \\
E & 35.5 & 1.64 & 3.67 & 2.57 & 2.96 & 0.01 & 41.1 & 7.41 & 3.72 & 3.05 & 3.04 & 0.04 & 4.60 \\ \bottomrule
\end{tabular}%
}
\end{table}

We further conduct ablation analysis on both random testset and biased testset. Besides the model ablations, we also ablate the dataset by removing the persona in the testset (w/o persona) to investigate whether such information is important. We report the ablation results in Table~\ref{tab:sparse-bob-ablation}, and we can find that:
1) Removing the persona information makes little difference to the random testset but has a significant influence on the performance of the biased testset.
2) The model ablation results on both testsets show that {${\mathbf{U}}$} contributes most to the perplexity, which is in agreement with the results on the PersonaChat dataset.
3) The significant performance gap between the initialization model E and the final model EDU$_{\text{BoB}}$ again proves the effectiveness of the stack-propagation framework.

\begin{table}[ht]
\small
\centering
\caption{Examples of response generation on PersonaChat. $\mathbb D$ and $\mathbb U$ are the decoder and regularizer, respectively.}
\label{tab:dense-case}
\begin{tabular}{@{}l|l@{}}
\toprule
\textbf{Persona} & My father used to work for home depot. I have a son who is in junior high school. \ \  \\
\textbf{Dialogue Query} & I recently started a job with the local high school. You have any children? \\
% \textbf{Ground-truth Response} & That's cool. My son is in junior high school. \\ 
\midrule
% \textbf{Transformer} & No, I do not. Do you? \\
\textbf{GDR} & Yes, I have two sons. \\
\textbf{LIC} & No children, but I am sure that my son is the next to be born soon. \\
\textbf{GPT-2} & No kids. I work at home depot so I'm busy. \\
\midrule
\multirow{2}{*}{\textbf{Stack-Propagation: BGB}} & $\mathbb D$: I do. My son is in high school. \\
 & $\mathbb U$: I do. My son is in high school. He's in junior high. \\
\multirow{2}{*}{\textbf{Stack-Propagation: GoG}} & $\mathbb D$: I have a son who is in junior high school. \\
 & $\mathbb U$: I have a son who is in junior high school. \\
\multirow{2}{*}{\textbf{Stack-Propagation: BoB}} & $\mathbb D$: Yes, I have a son. \\
 & $\mathbb U$: Yes, I have a son. He is in the 8th grade. \\ \midrule \midrule
\textbf{Persona} & I live in Alabama. I like tacos. I have one brother. I have three sisters. \\
\textbf{Dialogue Query} & Where do you live? I own a dog walking business. \\
% \textbf{Ground-truth Response} & Alabama. I love it here. \\ 
\midrule
% \textbf{Transformer} & I don't know. I also own my own business. \\
\textbf{GDR} & I live in Alabama. Where are you from? \\
\textbf{LIC} & I am from Alabama, but I live in Florida. \\
\textbf{GPT-2} & I live in New York. It is beautiful here. \\
\midrule
\multirow{2}{*}{\textbf{Stack-Propagation: BGB}} & $\mathbb D$: I do not live in a good neighborhood. I work in a law firm. \\
 & $\mathbb U$: I do not live in a good neighborhood. I work in a law firm. \\
\multirow{2}{*}{\textbf{Stack-Propagation: GoG}} & $\mathbb D$: I live in California with my 3 sisters \\
 & $\mathbb U$: I live in Alabama with my 3   sisters \\
\multirow{2}{*}{\textbf{Stack-Propagation: BoB}} & $\mathbb D$: Good, I also live in New York. \\
 & $\mathbb U$: Sounds good, I live in Alabama. \\ \midrule \midrule
\textbf{Persona} & I'm terrified of scorpions. I am employed by the US postal service. \\
\textbf{Dialogue Query} & My dad is an army soldier. \\
% \textbf{Ground-truth Response} & I thank him for his service. \\ 
\midrule
% \textbf{Transformer} & That is good, I know it is awsome \\
\textbf{GDR} & My dad was a cop and drowned when I was a kid. \\
\textbf{LIC} & Oh ok. What do you do for a living? \\
\textbf{GPT-2} & That sounds good! He was an army hero. \\
\midrule
\multirow{2}{*}{\textbf{Stack-Propagation: BGB}} & $\mathbb D$: I'm glad to hear that. I'm a postal worker. \\
 & $\mathbb U$: I'm glad to hear that. I'm a postal worker. \\
\multirow{2}{*}{\textbf{Stack-Propagation: GoG}} & $\mathbb D$: Thank you for serving our country. I am employed by the US postal service. \\
 & $\mathbb U$: Thanks for serving our country. I am employed by the US postal service. \\
\multirow{2}{*}{\textbf{Stack-Propagation: BoB}} & $\mathbb D$: That's cool. I'm a train driver. \\
 & $\mathbb U$: That's cool. I'm a bit of a Canadian who works for America. \\ \midrule \midrule
\textbf{Persona} & I'm fluent in English Spanish and French. I am an elementary school teacher. \\
\textbf{Dialogue Query} & I speak 2 languages, Spanish and English. \\
% \textbf{Ground-truth Response} & I speak those and French. \\
\midrule
% \textbf{Transformer} & Awesome. I am studying to be an elementary teacher. \\
\textbf{GDR} & Maybe I will take a special language class. \\
\textbf{LIC} & That's cool. what kind of music do you like? \\
\textbf{GPT-2} & My next language is Spanish. I hope to learn soon. \\
\midrule
\multirow{2}{*}{\textbf{Stack-Propagation: BGB}} & $\mathbb D$: I am fluent in Spanish and French. \\
 & $\mathbb U$: I am fluent in English, Spanish and French. \\
\multirow{2}{*}{\textbf{Stack-Propagation: GoG}} & $\mathbb D$: Cool! You must be really talented. \\
 & $\mathbb U$: Cool! You must be really talented. \\
\multirow{2}{*}{\textbf{Stack-Propagation: BoB}} & $\mathbb D$: That is awesome! I speak them, too. \\
 & $\mathbb U$: That is awesome! I speak both English and French, and Spanish too. \\ \bottomrule
\end{tabular}
\end{table}

\begin{table}[ht]
\centering
\small
\captionsetup{width=.94\textwidth}
\caption{Examples of consistency understanding capability, measured by the perplexity. The \textit{support} denotes the negative example that is similar to the persona in the language inference datasets. Larger perplexity indicates less generation probability.}
\label{tab:dense-case-support}
\begin{tabular}{@{}l|l@{}}
\toprule
\textbf{Persona} & I have a son who is in junior high school. \\
\textbf{Dialogue Query} & I recently started a job with the local high school. You have any children? \\
\textbf{Ground-truth Response} & That's cool. My son is in junior high school. \\ \midrule
\multirow{2}{*}{\textbf{Stack-Propagation: BGB}} & Response: I do. My son is in high school. He's in school. [PPL=16.8] \\
 & Support:  I've got a son $\Rightarrow$ I don't have kids [PPL=136.2] \\ \midrule
\multirow{2}{*}{\textbf{Stack-Propagation: GoG}} & Response: I have a son who is in junior high school. [PPL=15.2] \\
 & Support:  I've got a son $\Rightarrow$ I don't have kids [PPL=101.6] \\ \midrule
\multirow{2}{*}{\textbf{Stack-Propagation: BoB}} & Response: Yes, I have a son. He is in the 8th grade. [PPL=13.1] \\
 & Support: I've got a son $\Rightarrow$ I don't have kids [PPL=158.3] \\ \midrule \midrule
\textbf{Persona} & I live in Alabama. I like tacos. I have one brother. I have three sisters. \\
\textbf{Dialogue Query} & Where do you live? I own a dog walking business. \\
\textbf{Ground-truth Response} & Alabama. I love it here. \\
\midrule
\multirow{2}{*}{\textbf{Stack-Propagation: BGB}} & Response: I do not live in a good neighborhood. I work in a law firm. [PPL=18.8] \\
 & Support:  I'm native to Alabama $\Rightarrow$ I'm a native born Texan [PPL=113.6] \\ \midrule
\multirow{2}{*}{\textbf{Stack-Propagation: GoG}} & Response: I live in Alabama with my 3   sisters [PPL=14.7]\\
 & Support:  I'm native to Alabama $\Rightarrow$ I'm a native born Texan [PPL=92.4] \\ \midrule
\multirow{2}{*}{\textbf{Stack-Propagation: BoB}} & Response: Sounds good, I live in Alabama. [PPL=10.0] \\
 & Support:  I'm native to Alabama $\Rightarrow$ I'm a native born Texan [PPL=123.7] \\ \midrule \midrule
\textbf{Persona} & I am employed by the US postal service. \\
\textbf{Dialogue Query} & My dad is an army soldier. \\
\textbf{Ground-truth Response} & I thank him for his service. \\
\midrule
\multirow{2}{*}{\textbf{Stack-Propagation: BGB}} & Response: I'm glad to hear that. I'm a postal worker. [PPL=20.9] \\
 & Support:  You are a TI employee $\Rightarrow$ Not a TI employee, you work at Burger King. [PPL=129.4] \\ \midrule
\multirow{2}{*}{\textbf{Stack-Propagation: GoG}} & Response: Thanks for serving our country. I am employed by the US postal service. [PPL=14.5] \\
 & Support: You are a TI employee $\Rightarrow$ Not a TI employee, you work at Burger King. [PPL=96.3] \\ \midrule
\multirow{2}{*}{\textbf{Stack-Propagation: BoB}} & Response: That's cool. I'm a bit of a Canadian who works for America. [PPL=14.9] \\
 & Support:  You are a TI employee $\Rightarrow$ Not a TI employee, you work at Burger King. [PPL=116.2] \\
\bottomrule
\end{tabular}
\end{table}

\subsection{Case Study}
\label{sec:exp_cases}
To better understand how the proposed stack-propagation framework works, we conduct some case studies.
We first show some generated responses from different models on the PersonaChat dataset in Table~\ref{tab:dense-case}. For the EDU framework, we present the generated responses from both decoders, including decoder $\mathbb D$ and regularizer $\mathbb U$, to have a better view of how the stack-propagation framework works.
From these cases we can see that:
1) All the presented methods are capable of effectively leveraging persona information, but the baseline models are usually mimicking the personas, sometimes even leading to contradictions. Taking the case ``No kids. I work at home depot so I'm busy.'' from the GPT-2 as an example: it copies words from the persona ``My father used to work for home depo'' but misuses the information to generate the response. While the true information ``I have a son who is in junior high school.'' is neglected. Such phenomena indicate that even the pre-trained language models still lack a good understanding of consistency.
2) In the stack-propagation EDU framework, the responses from $\mathbb D$ and $\mathbb U$ are similar in their skeletons, but the details are different in most cases. These differences indicate that the regularizer $\mathbb U$ has obtained some consistency understanding capability.
3) All the final responses (response from $\mathbb U$) of the stack-propagation framework are appropriate not only in content but also in persona consistency, demonstrating the effectiveness of the EDU framework for considering the generation of responses and the understanding of persona consistency simultaneously.

We further present some examples together with their perplexities in Table~\ref{tab:dense-case-support} to showcase how the stack-propagation framework understands consistency.
Besides the perplexities of the responses, we also showcase the regularizer $\mathbb U$'s perplexities of negative examples in the language inference dataset. Here the premises of the negative examples are similar to the personas.
From the shown examples, we can see that:
1) The perplexities of most generated responses are less than 20, showing that the EDU models are quite confident about their generation results.
2) The $\mathbb U$'s perplexities for negative examples are very large, indicating a very small generation probability.
3) The perplexity differences between generated responses and negative examples can be consistently observed in three initialization schemes, showing the generalization of leveraging $\mathbb U$ as a regularizer for the response generation.

\section{Conclusion}
In this work, we propose a stack-propagation framework named EDU for the personalized dialogue generation task, and the EDU framework is especially effective under low-resource scenarios. We formulate the personalized dialogue generation task as a response generation task regularized by a consistency understanding task. In this way, we can leverage abundant non-dialogue data to train both sub-tasks. Specifically, we introduce a stacked pipeline of transformer blocks consisting of an encoder, a decoder, and a decoder-based regularizer. We explore different initialization schemes with both the auto-regressive language models and masked language models. Evaluations are carried out on a persona-dense dataset and further validated on a persona-sparse dataset. Experimental results show that the stack-propagation framework can generate consistent responses and establish a stable performance under low-resource scenarios, significantly outperforming strong baseline models under both automatic and human evaluations.

This work shows how to leverage existing pre-trained language models to achieve significantly better performance on the personalized dialogue generation task. We believe leveraging understanding as a regularization of generation can be extended to other generation tasks based on the stack-propagation framework. We will continue to explore better architectures that leverage pre-trained language models for the unique challenges in dialogue tasks.

\bibliographystyle{ACM-Reference-Format}
\bibliography{main}

% \appendix


\end{document}